\documentclass[11pt,a4paper]{article}
\usepackage{times,latexsym}
\usepackage{url}
\usepackage[T1]{fontenc}
\usepackage{algorithm}
\usepackage{algorithmic}
\usepackage{amsmath}
\usepackage{amssymb}
\usepackage{enumerate}
\usepackage{booktabs}

\usepackage[acceptedWithA]{tacl2021v1}
\usepackage{tacl2021v1}
\usepackage{xcolor,graphicx,subfig,lipsum,colortbl,multirow}

\usepackage{enumitem}

\usepackage{xspace,mfirstuc,tabulary}

\newif\iftaclinstructions
\taclinstructionsfalse 
\iftaclinstructions
\renewcommand{\confidential}{}
\renewcommand{\anonsubtext}{(No author info supplied here, for consistency with
TACL-submission anonymization requirements)}
\newcommand{\instr}
\fi

\iftaclpubformat 

\else

\fi

\title{Beyond Majority Voting: Agreement-Based Clustering to Model Annotator Perspectives in Subjective NLP Tasks}

\author{
  \textbf{Tadesse Destaw Belay\textsuperscript{1}},
  \textbf{Ibrahim Said Ahmad\textsuperscript{2}},
  \textbf{Idris Abdulmumin\textsuperscript{3}},
  \textbf{Abinew Ali Ayele\textsuperscript{4}},\\
  \textbf{Alexander Gelbukh\textsuperscript{1}},
  \textbf{Eusebio Ricárdez-Vázquez\textsuperscript{1}},
  \textbf{Olga Kolesnikova\textsuperscript{1}},\\
  \textbf{Shamsuddeen Hassan Muhammad\textsuperscript{2,5}},
  \textbf{Seid Muhie Yimam\textsuperscript{6}}
  \\
$^{1}$Instituto Politécnico Nacional,
$^{2}$University of Wisconsin–Stevens Point,\\
$^{3}$University of Pretoria,
$^{4}$Bahir Dar University,\\
$^{5}$Imperial College London,
$^{6}$University of Hamburg \\
{\small\texttt{
tadesseit@gmail.com}}}
\date{}

\begin{document}
\maketitle
\begin{abstract}

Disagreement in annotation is a common phenomenon in the development of NLP datasets and serves as a valuable source of insight. While majority voting remains the dominant strategy for aggregating labels, recent work has explored modeling individual annotators to preserve their perspectives.  
However, modeling each annotator is resource-intensive and remains underexplored across various NLP tasks. We propose an agreement-based clustering technique to model the disagreement between the annotators. We conduct comprehensive experiments in 40 datasets in 18 typologically diverse languages, covering three subjective NLP tasks: sentiment analysis, emotion classification, and hate speech detection. We evaluate four aggregation approaches: majority vote, ensemble, multi-label, and multitask. The results demonstrate that agreement-based clustering can leverage the full spectrum of annotator perspectives and significantly enhance classification performance in subjective NLP tasks compared to majority voting and individual annotator modeling. Regarding the aggregation approach, the multi-label and multitask approaches are better for modeling clustered annotators than an ensemble and model majority vote. The dataset is publicly available in GitHub: \url{https://github.com/Tadesse-Destaw/Beyond-Majority-Voting}.
 
\end{abstract}

\section{Introduction}
In supervised machine learning, using multiple annotators to label datasets is a common strategy to improve the quality of training and evaluation data for downstream natural language processing (NLP) tasks \cite{Cabitza_2023}. However, annotators' disagreements frequently arise during the annotation process. These disagreements stem not only from random errors but also from systematic differences in task interpretation and understanding. Specifically, sociodemographic factors, such as age, gender, race, educational status, political stance, and living experience, can significantly influence how individuals interpret subjective annotation tasks \cite{luo-etal-2020-detecting,zhang2022examining,beck-etal-2024-sensitivity}. Consequently, annotators often approach the same text from different perspectives, leading to divergent judgments shaped by their personal and cultural connotations \cite{Wan-Kim-Kang-2023,lee-etal-2024-exploring-cross}.

Current approaches for handling annotator disagreement fall into four categories: (1) Aggregating annotations using majority vote, an extra expert as a judge, or Bayesian methods to get a single ground truth label \cite{paun-etal-2018}; (2) Treating each annotator’s label as potentially valid while filtering uncertain items due to disagreement  \cite{wang-plank-2023-actor}; 
(3) Training models directly on the raw annotations without label aggregation \cite{daval-frerot-weis-2020-wmd}; and (4) Leveraging both hard and soft labels from annotations during model training \cite{federated-2024}.

Majority voting is the most common practice to decide the final annotation label \cite{Uma-survey-2022,xu-etal-2024-leveraging}. However, this approach removes genuine disagreement between annotators, thereby marginalizing minority viewpoints that could offer valuable insights \cite{leonardelli-etal-2023-semeval,rizzi-etal-2025-bunch}. Although adequate for objective tasks like part-of-speech tagging (where even a single annotator label may suffice), majority voting poses limitations in subjective annotation tasks, where establishing a single ground truth is inherently challenging \cite{khurana2024crowdcalibrator}. 

In recent years, the practice of only considering the majority vote has been criticized \cite{sandri-etal-2023-dont}. Research has started to advocate for better ways to deal with disagreements between annotators and to preserve all interpretations of annotators rather than eliminate the minority point of view \cite{Uma-survey-2022, davani_dealing_2022, federated-2024}. Considering each annotator's perspective is an emerging research direction in computational subjective tasks \cite{yin2023annobert}, for instance, modeling annotation disagreement, as seen in shared tasks and workshops focusing on disagreement in subjective tasks \cite{leonardelli-etal-2023-semeval,comedi-ws-2025-1}. 

A common approach to capturing annotation disagreement while preserving individual perspectives is to model each annotator separately \cite{davani_dealing_2022,federated-2024}. However, training a model for each annotator is computationally resource-intensive, especially when the number of annotators is large and the annotation is skewed among annotators. In this work, we propose an agreement-based clustering technique that groups annotators by their agreement patterns while maintaining individual perspectives. 
Our contributions are:
\begin{itemize}
\item \textbf{An agreement-based clustering framework} that automatically groups annotators by their agreement patterns, preserving diverse perspectives while significantly reducing computational overhead compared to per-annotator modeling;
\item \textbf{Large-scale empirical validation} across 40 multilingual datasets covering three challenging subjective tasks (sentiment analysis, emotion classification, and hate speech detection), showing consistent improvements over individual annotator modeling approaches;
\item \textbf{A comprehensive comparison} of aggregation strategies (ensemble, multi-label, and multitask), establishing baselines against majority voting.
\end{itemize}

\section{Literature Review}
In this section, we review the literature relevant to our work, focusing on subjective tasks, the sources of disagreement, and approaches to modeling annotators disagreement in subjective NLP tasks.

\subsection{Subjective NLP Tasks}
Subjective NLP tasks incorporate diverse answers during annotation due to the differences between the sociodemographic backgrounds of annotators \cite{rottger-etal-2022-two}. Finding a single true label in such tasks using majority vote can lead to biased results \cite{Uma-survey-2022}, and it is crucial to incorporate each perspective of the annotator. Examples of subjective tasks include sentiment analysis \cite{muhammad-etal-2022-naijasenti,muhammad-etal-2023-afrisenti,978-3-032-12993-2_29,afriyie2026sentiment}, hate speech \cite{kapil-ekbal-2024-survey,beck-etal-2024-sensitivity,muhammad2025afrihate}, abusive speech \cite{g-etal-2025-overview}, humor and sarcasm identification \cite{simpson-etal-2019-predicting}, toxicity \cite{van-aken-etal-2018-challenges}, ironic content \cite{frenda-etal-2023-epic}, good or bad \cite{martinez2018overview} and emotion classification \cite{muhammad-etal-2023-afrisenti,muhammad-etal-2025-semeval}. Obtaining high-quality and reliable annotations in such subjective tasks is challenging. It is common to see low agreement among annotators in such tasks, such as GoEmotions emotion data \cite{demszky-etal-2020-goemotions} has 27\% Cohen's kappa (considered low), and HateXplain \cite{hatexplain}  hate speech data has 46\%  (considered moderate agreement) \cite{kappa}.

\subsection{Sources of Disagreement}
Disagreement in the annotations of NLP datasets refers to the absence of a single ground truth label, often arising from genuine differences in annotator interpretation \cite{rottger-etal-2022-two,braun2024beg}. Traditionally viewed as noise, such disagreement is now increasingly recognized as a valuable source of information \cite{fell2021mining}. While all forms of disagreement contribute to label uncertainty, their underlying reasons may differ. Broadly, three perspectives explain the origins of annotation disagreement: (1) issues related to annotation design, such as unclear guidelines, poorly defined label spaces, lack of contextual information, or low-quality annotator performance  \cite{denton2021whose,parmar-etal-2023-dont}; (2) inherent ambiguity in the text itself \cite{Uma-survey-2022}; and (3) variation in annotators’ perspectives which is influenced by sociodemographic factors \cite{jiang-marneffe-2022-investigating,rethinkingemotion2024}.

\subsection{Modeling Annotation Disagreement}

Leveraging annotation disagreement during model training has been demonstrated to serve as a valuable learning signal. \citet{hayat2022modeling} implemented multitask-based modeling of each annotator separately, which yields better performance than the traditional majority-vote approach. \citet{federated-2024} proposed federated learning that builds a global model from the client annotator embedding models. \citet{mokhberian-etal-2024-capturing} proposed an annotator-aware representation for texts (AART) that combines text with annotator embeddings. Several works (e.g., \citet{davani_dealing_2022,federated-2024,xu-etal-2024-leveraging}) have proposed ensemble-based approaches that model each annotator individually. 

However, training a separate model for each annotator can be computationally expensive and often impractical in real-world settings. For instance, the GoEmotions \cite{demszky-etal-2020-goemotions} emotion dataset annotated by over 82 annotators, a sentiment analysis dataset by \citet{diaz2020biases} involved more than 1400 annotators, and the hate speech dataset by \citet{Jigsaw-2019} was constructed with contributions from over 8000 annotators. Moreover, the number of examples annotated per individual varies considerably. In the GoEmotions dataset, for instance, annotator contributions range from as few as 3 to as many as 4800 instances.

Previous studies have examined annotation disagreement at various levels. 
\citet{Matteo-community-based-bayesian} proposed a probabilistic Bayesian model that jointly learns latent community profiles of crowd workers and estimates both worker reliability and true labels by clustering workers into communities.
Similarly, \citet{Himabindu-2015} proposed a hierarchical Bayesian framework to evaluate the quality of individual annotators and to identify the true labels of items. Their work also diagnoses the types of errors that annotators tend to make and the common characteristics of items on which such errors occur, including clustering annotators who share similar attributes. \citet{weerasooriya-etal-2022-improving} introduced a graphical model that enhances the quality of annotator labels by using clustering to strengthen the signal of noisy data. Finally, \citet{jury-learning-gordon} addressed annotators disagreement through jury learning, which involves selecting a juror from each representative group. This method models each annotator in a dataset and requires sociodemographic metadata such as self-identified gender, race, education, political affiliation, age, parental status, and religiosity to produce a joint jury prediction for classifying unseen examples.

However, these studies primarily focus on capturing the quality and error patterns of individual evaluators during the annotation process \cite{weerasooriya-etal-2022-improving,jury-learning-gordon}, after annotation when detail annotator attributes are available \cite{Himabindu-2015}, or by requiring an additional expert (gold) judgments from highly trained editorial annotators to serve as ground truth for evaluating regular annotators \cite{Matteo-community-based-bayesian}.
In contrast, our work focuses on the post-annotation stage and aims to model the annotators’ perspectives (opinions) to estimate the true labels more accurately before influenced by majority voting. Our approach is designed for already annotated datasets where only the anonymous annotator IDs are available.

Most similar to our work (\citet{fell2021mining,parting-crowds-2016,lo2023hierarchical}) proposed clustering or grouping the annotations based on the agreement of annotators. However, these clustering approaches relied on the annotators' metadata, such as culture, demographics, and ethnic backgrounds. This clustering approach works well when the annotators' sociodemographic metadata is available. 

Based on the limitations of the approaches implemented in the previous works \cite{daval-frerot-weis-2020-wmd,fell2021mining,davani_dealing_2022,federated-2024}, we propose clustering annotators based on their annotation agreements to overcome over-modeling of annotators; more details are in Section  §\ref{agree-based}. In addition, the previous exploration of subjective tasks solely relies on the English language; we covered various low-resource languages, annotation taxonomies, data sources, and subjective tasks. Finally, we evaluate modeling individual annotators versus agreement-based clustered annotators using ensemble, multi-label, and multitask aggregation approaches, with the majority vote as a baseline.

\section{Agreement-based Annotator Clustering}\label{agree-based}

\begin{figure*}[h]
 \centering
  \subfloat[Annotators (their IDs) and its agreement]{\includegraphics[width=.49\textwidth]{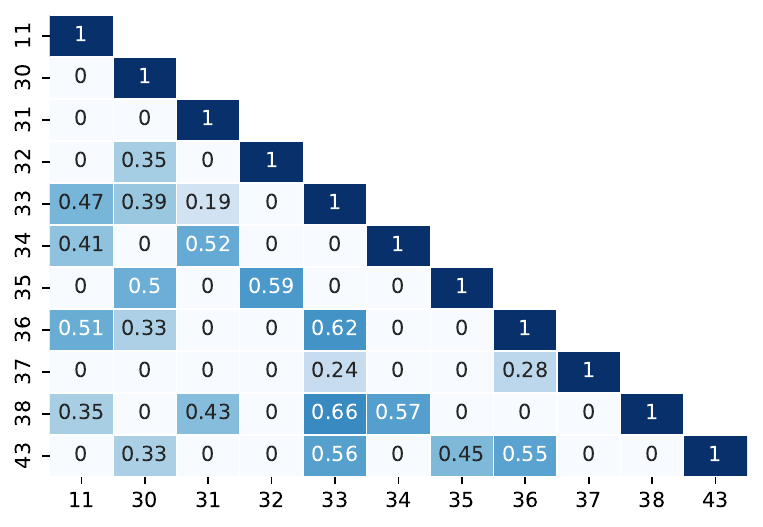}}
  \subfloat[Clustering the 11 annotators into $C=3$]{\includegraphics[width=.49\textwidth]{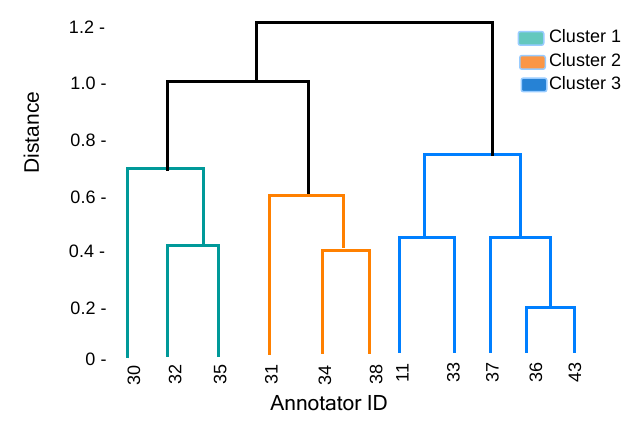}}
  \caption{Pairwise Agreement between annotators. This agreement score is used to group annotators and determine which annotator belongs to which cluster. The example is from the Nigerian Pidgin (\textit{pcm}) language sentiment dataset — eleven annotators participated, with each instance annotated by a minimum of $3$, and the annotators are clustered as cluster $C1= 30,32,35$, $C2 =31,34,38$, and $C3 = 11, 33, 36,37, 43$.}
  \label{fig:agreement}
\end{figure*}

A common approach for modeling disagreement is to model each annotator separately \cite{davani_dealing_2022} or a sample of annotators \cite{yin2023annobert}.  While training a model for each annotator has the advantage of preserving individual perspectives, it has two drawbacks: (1) when many annotators participate, annotation distributions become highly skewed; and (2) training one model per annotator is computationally intensive and expensive. Instead of modeling each annotator separately, clustering annotators based on their agreements could address this issue.

The clustering approach is motivated by the premise that if two annotators independently assign the same label, such as \textit{hate} in a hate speech annotation task, they demonstrate agreement on that specific instance. This agreement suggests a shared perspective between the annotators, justifying their inclusion within the same cluster. We proposed an agreement-based clustering of annotators into a predefined number of clusters to preserve the perspectives of annotators while clustering. The approach is explained in the following conditions:

~\\ \textbf{Condition 1:} For a dataset, if the number of annotation per instance ($n$) and the total number of participating annotators ($N$) is equal, we treat each annotator as a distinct cluster. For example, if three annotators annotate each instance in a dataset and there are three participating annotators, then the data naturally form three clusters, with each annotator forming a cluster. In this case, the agreement-based clustering approach is not applied, and the number of clusters $C$ equals the number of annotators $N$.

~\\ \textbf{Condition 2:} For a dataset annotated by $N$ annotators with variable annotation coverage: 
\begin{enumerate}
\item \textbf{Label matrix construction}
Let $\mathcal{X} = \{x_1, \ldots, x_M\}$ be the set of data instances. Suppose there are $N$ annotators, $\mathcal{A} = \{a_1, \ldots, a_N\}$. Annotator $a_i$ labels some instances $x_j$ as $l_{i,j}$. We construct an annotation matrix $L \in \mathbb{R}^{N \times M}$, where entry $L_{i,j}$ is the label from $a_i$ for $x_j$. If $a_i$ did not label $x_j$, the entry is left empty.
\item \textbf{Annotator similarity matrix}
For each pair of annotators $(a_i, a_k)$:
\begin{itemize}[leftmargin=0pt]
    \item Identify the set of co-annotated instances:
    $S_{i,k} = \{ x_j \mid l_{i,j} \ \text{and} \ l_{k,j} \ \text{exist} \}$
    \item Compute an agreement score (Cohen's kappa for multiclass and Jaccard similarity for multi-label comparing sets of labels):

$\text{sim}(i,k)=c(\{l_{i,j}\mid x_j\in S_{i,k}\},\{l_{k,j}\mid x_j\in S_{i,k}\})$
    where $c(\cdot, \cdot)$ is the agreement function.
\end{itemize}
All $\text{sim}(i, k)$ values form the similarity matrix $A \in \mathbb{R}^{N \times N}$. We convert to a distance matrix:$D_{i,k} = 1 - \text{sim}(i, k)$

\item  \textbf{Annotator clustering}
Apply a clustering algorithm (e.g., k-means clustering) using $D$ as the distance matrix. The number of clusters $C$ is set as: 
$C = \min_j \left| \{ i : l_{i,j} \ \text{exists} \} \right|,$ ensuring each instance is represented by at least one cluster.

The result is a set of clusters $\mathcal{C}_1, \ldots, \mathcal{C}_C$, with $\mathcal{C}_c \subseteq \mathcal{A}$.
\item \textbf{Cluster-level label aggregation}
For each instance $x_j$ and cluster $\mathcal{C}_c$:
\begin{itemize}[leftmargin=0pt]
    \item Collect all labels for $x_j$ from annotators in $\mathcal{C}_c$.
    \item If a majority label is available between, assign it as the cluster label for $x_j$.
    \item \textbf{If there is a tie or multi-label:}
    \begin{itemize}
        \item If the task is multi-label, such as the multi-label emotion dataset, assign all available labels.
        \item  Otherwise, find a third or fifth annotator (adjudicator) for the specific tied instance as a tie-breaker and take the majority vote.
    \end{itemize}
\end{itemize}

So for each $x_j$, we have $C$ cluster labels - a label(s) after aggregations of annotators.

\end{enumerate}

\paragraph{Number of Clusters} We consider conditions to decide the number of clusters. 1) The minimum number of annotators can be found in any annotation condition, including instances that have more annotators. For example, if the minimum number of annotators per instance in a dataset is 3, the cluster can be 3; this covers cases with more annotators, and the number of clusters can be determined based on the availability of computational resources and data factors. 
2) At the end, the results of the disagreement modeling approach is evaluated against the majority-voted gold labels. To decide the instance $ x_j$'s label(s) within the cluster, the number of clusters is preferably kept odd (e.g., 3, 5, ...) because an odd number of clusters eliminates the possibility of ties, as detailed in Algorithm \ref{alg:clustering}.

\begin{algorithm}[h]
\small
\caption{Annotators Clustering}
\label{alg:clustering}
\begin{algorithmic}[1]
\REQUIRE Label matrix $L \in \mathbb{R}^{N \times M}$
\REQUIRE Agreement function $c(\cdot, \cdot) orj(\cdot, \cdot)$
\ENSURE Annotator clusters based on agreement
\STATE Compute similarity matrix $A$ from $L$ using $c$
\STATE Compute distance matrix $D = 1 - A$
\STATE Number of clusters $C = \min_j\left|\{i:l_{i,j}\ \text{exists}\}\right|$
\STATE Cluster annotators into $\mathcal{C}_1, \ldots, \mathcal{C}_C$ using $D$
\FOR{each instance $x_j$}
    \FOR{each cluster $\mathcal{C}_c$}
        \STATE Aggregate labels for $x_j$ from $\mathcal{C}_c$ members
        \STATE Assign cluster label for $x_j$
    \ENDFOR
\ENDFOR
\STATE \textbf{return} Annotator clusters
\end{algorithmic}
\end{algorithm}

\begin{figure*}
    \centering
    \includegraphics[width=0.90\textwidth]{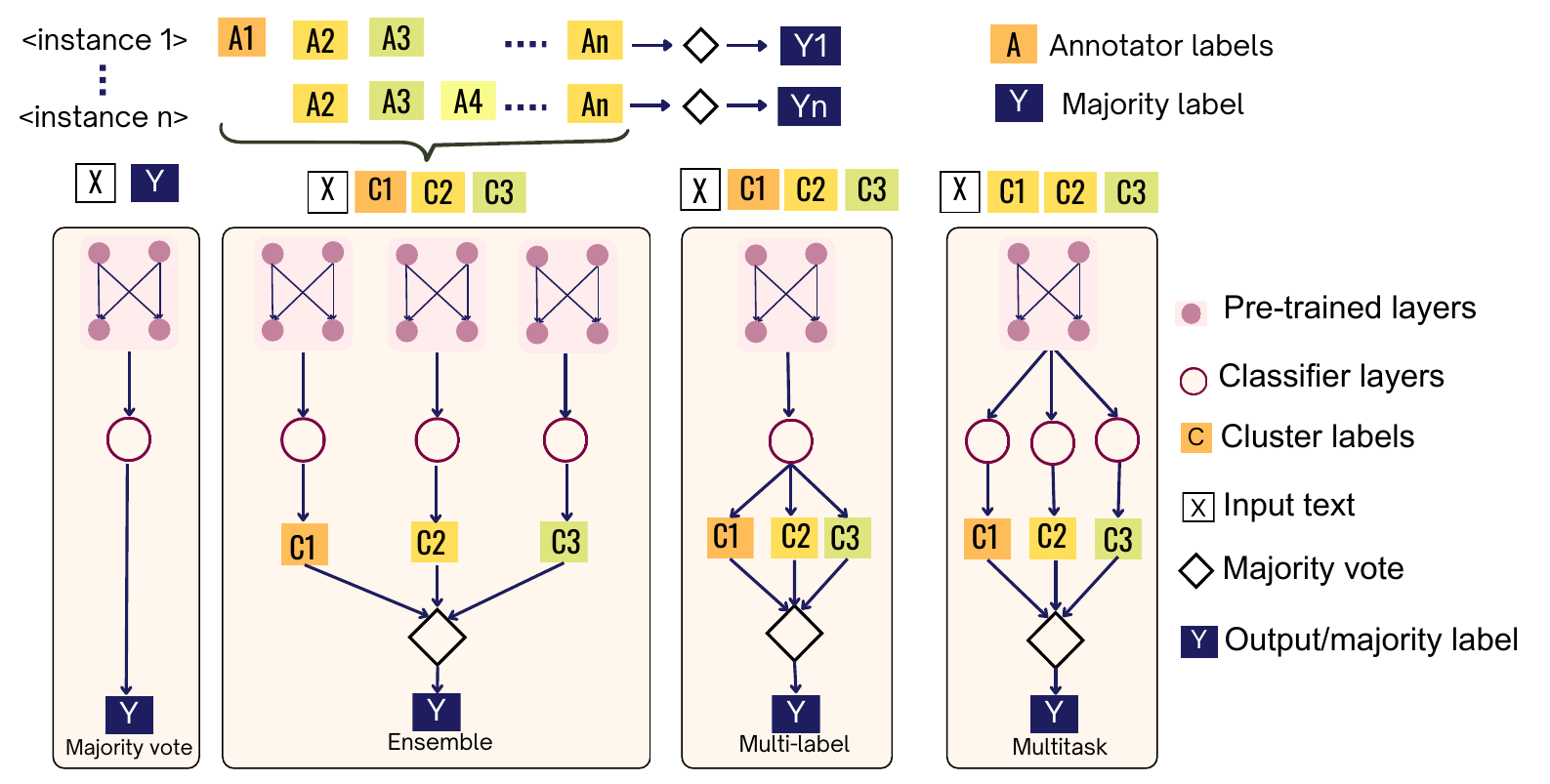}
    \caption{Overview of multi-annotator modeling architectures. Our new contribution is in the clustering of annotators before modeling. In the figure, $N$ annotators, $A1, A2, A3, ... An$, are clustered into $C$ clusters based on annotator label agreement. For visualization, we limit the number of clusters to three.}
    \label{fig:method}
\end{figure*}

A high label agreement $c(i, j)$ indicates that annotators $i, j$ tend to give similar labels on the items (texts). The lower distance indicates higher agreement based on their distances $D$ to one another; annotators and their labels are clustered into a predefined number of clusters using $k-means$ clustering. 
After clustering, we evaluate three commonly used \textbf{aggregation approaches} in addition to the majority vote baseline: ensemble, multi-label, and multitask learning. These methods operate on non-aggregated annotator labels, preserving the individual perspectives rather than aggregating them into a single label, as is done in the majority vote. Figure \ref{fig:method} illustrates the schematic differences between these approaches.

\subsection{Majority Vote}
The majority vote is the most commonly used baseline, which aggregates annotations without considering annotator-specific information. It involves training a single-task classifier to predict the aggregated label for each instance, where the gold label is determined by majority voting. We use the majority vote as a baseline for comparison with the proposed approaches.

\subsection{Ensemble Approach}

This approach trains a separate model for each annotator to predict their individual labels. During inference, the predictions of these individual models are aggregated via majority vote to produce the final label \cite{akhtar2020modeling}. In our work, we extend this idea to both annotator-level and cluster-based ensembles. We train a model for individual annotators as well as clustered subsets, and evaluate their outputs independently.

\subsection{Multi-label Approach}
This approach treats each annotation as an individual label and formulates the problem as a multi-label classification task \cite{davani_dealing_2022}. The model incorporates a fully connected layer to project the input representation into a vector of annotator-specific label predictions, followed by a sigmoid activation to obtain independent probability scores for each annotator. 

\subsection{Multitask Approach} 
The multitask approach models each annotator or cluster as a separate classification task. All tasks share a common encoder to generate sentence representations, while each task has its own task-specific fully connected layer followed by a softmax activation. In contrast to the multi-label approach, multitask learning explicitly trains a separate output layer for each task \cite{liu-etal-2019-multi-lingual}. 

\section{Data} \label{sec:data}
This section provides an overview of the datasets used in this study. Although many publicly available datasets exist, annotator-level datasets remain scarce, particularly for languages other than English. For this study, we selected recently available datasets for three highly subjective NLP tasks: hate speech detection, sentiment analysis, and emotion classification. These datasets include only anonymous annotator identifiers (ID numbers), which enable clustering based on annotation behavior. The summary of the dataset is presented in Table \ref{tab:dataset}.

\begin{table*}[h!]
\centering
\resizebox{\textwidth}{!}
{\begin{tabular}{llccccp{1cm}p{1cm}rrr}
\hline
\textbf{Task} &\textbf{Language (ISO code)} &\textbf{Train}&  \textbf{Dev}&  \textbf{Test}& \textbf{Total}&  \textbf{\#Anno /inst.}&\textbf{\#Total Anno.}& \textbf{Labels distribution in \%} & \textbf{\%full agr.}& \textbf{C. Kappa}\\
\hline
\multirow{13}{*}{\rotatebox[origin=c]{90}{\textbf{Sentiment analysis}}}
& Amharic (\textbf{\texttt{amh}}) & 1,300 & 200 & 500 & 2,000 & 3 &3 & Pos (22), Neg (67), Neu (11) &63.55& 53.65 \\
& Moroccan Arabic (\textbf{\texttt{ary}}) & 2,971 & 458 & 1,142 & 4,571 & 3 &9 & Pos (30), Neg (34), Neu (36) &52.48 & 52.55 \\
& Hausa (\textbf{\texttt{hau}}) &15,360 & 2,364 & 5,907 & 23,631 & 3&3& Pos (30), Neg (29), Neu (42)&61.46 & 61.07 \\
& Igbo (\textbf{\texttt{ibo}})  & 19,215 & 2,957 & 7,391 & 29,563 & 3&3 & Pos (25), Neg (22), Neu (53)&62.86 & 59.66 \\
& Kinyarwanda (\textbf{\texttt{kin}})  & 2,615 & 403 & 1,006 & 4,024 & 3&3& Pos (26), Neg (33), Neu (40)&38.10 & 37.96 \\
& Oromo (\textbf{\texttt{orm}})  & 1,749 & 270 & 672 & 2,691 & 3&3& Pos (24), Neg (35), Neu (41)& 39.58& 35.11 \\
& Nigerian Pidgin  (\textbf{\texttt{pcm}}) & 11,207 & 1,725 & 4,310 & 17,242 & 3&11& Pos (34), Neg (55), Neu (12)&50.78 & 46.36 \\
&Portuguese-MZ (\textbf{\texttt{ptMZ}}) & 5,667 & 872 & 2,180 & 8,719 & 3&6& Pos (16), Neg (18), Neu (66)&45.83 & 33.82 \\
& Tigrinya (\textbf{\texttt{tir}}) & 1,560 & 240 & 600 & 2,400 & 3&3& Pos (29), Neg (49), Neu (21)&65.83 & 61.81 \\
& Xitsonga (\textbf{\texttt{tso}})  & 662 & 111 & 331 & 1,104 & 3&3 & Pos (47), Neg (35), Neu (18)&43.12 & 43.19 \\
& Twi (\textbf{\texttt{twi}})  & 3,043 & 469 & 1,171 & 4,683 & 3&3 & Pos (47), Neg (38), Neu (15)&49.54 & 47.16 \\
& Yorùbá (\textbf{\texttt{yor}})  & 17,807 & 2,740 & 6,849 & 27,396 & 3&4& Pos (34), Neg (17), Neu (49)&63.00 & 60.19 \\

\hline
\multirow{15}{*}{\rotatebox[origin=c]{90}{\textbf{Emotion classification}}}
& Amharic (\textbf{\texttt{amh}})  & 3,844 & 593 & 1,478 & 5,915 & 5&5& 26,  28,  2,  12,  16,  4,  12&11.11& 39.52 \\
& English (\textbf{\texttt{eng}})  & 43,410 & 5,426 & 5,427 & 54,263 & 3.58&82& 
3, 0.5, 0.5, 8, 1.5, 2.5, 6&31.98& 29.40 \\ 

& Moroccan Arabic (\textbf{\texttt{ary}})  & 1,746 & 270 & 671 & 2,687 & 3&3 & 21, 4, 7, 17, 14, 14, 25 &26.87 & 41.63 \\
& Hausa (\textbf{\texttt{hau}})  & 2,327 & 359 & 895 & 3,581 & 5.03&7 & 16, 12, 12, 12, 25, 13, 9 &23.21 & 54.13 \\
& Igbo (\textbf{\texttt{ibo}})  & 3,121 & 481 & 1,201 & 4,803 & 3&3& 19, 18, 7, 16, 17, 3, 21 &32.50 & 46.66 \\
& Kinyarwanda (\textbf{\texttt{kin}})  & 2,656 & 410 & 1,023 & 4,089 & 3&3& 17, 5, 5, 16, 24, 4, 28 &50.55 & 63.27 \\
& Oromo (\textbf{\texttt{orm}}) & 3,499 & 539 & 1,346 & 5,384 & 3 &6& 17, 16, 3, 31, 8, 4, 22 &43.59 &53.58 \\
& Nigerian Pidgin  (\textbf{\texttt{pcm}})  & 4,041 & 623 & 1,554 & 6,218 & 3&3 & 6, 35, 8, 11, 18, 18, 3 &13.48 & 43.03 \\
&Portuguese-MZ (\textbf{\texttt{ptMZ}})  & 1,243 & 208 & 622 & 2,073 & 4&4& 7, 5, 9, 16, 22, 0, 41 &30.05 & 20.82 \\
& Somali (\textbf{\texttt{som}})& 3,229 & 498 & 1,242 & 4,969 & 3&6& 10, 12, 8, 17, 12, 5, 36 &30.83 & 42.53 \\
& Swahili (\textbf{\texttt{swa}})  & 3,583 & 552 & 1,379 & 5,514 & 3 &3& 9, 7, 3, 13, 10, 16, 43 &14.51 & 19.97 \\
& Tigrinya (\textbf{\texttt{tir}})  & 3,583 & 554 & 1,380 & 5,523 & 3&6& 13, 31, 3, 10, 14, 9, 19 &24.21 & 41.50 \\
& isiXhosa (\textbf{\texttt{xho}}) & 1,365 & 228 & 682 & 2,275 & 3&3& 4, 0.5, 2, 28, 38, 15, 13 &39.74 &  41.15 \\
& Yorùbá (\textbf{\texttt{yor}}) & 3,242 & 500 & 1,247 & 4,989 & 3&3 & 6, 3, 3, 9, 27, 8, 44 &33.09 & 33.79 \\
& isiZulu (\textbf{\texttt{zul}})  & 1,899 & 293 & 730 & 2,922 & 3&3 & 8, 3, 3, 5, 19, 8, 54  &46.17& 40.05 \\

\hline
\multirow{12}{*}{\rotatebox[origin=c]{90}{\textbf{Hate speech classification}}}
& Amharic (\textbf{\texttt{amh}})  & 3,132 & 482 & 1,205 & 4,819 & 3&11& Hate (45), Abuse (28), Neu (27)&53.46 & 59.67 \\
& English (\textbf{\texttt{eng}})  & 22,124 & -- & 5,531 & 27,655 & 3.13&18& Hate (2,599), non-hate (24,977)&75.38 & 28.00 \\
& Moroccan Arabic (\textbf{\texttt{ary}})  & 2,704 & 416 & 1,040 & 4,160 & 3&3& Hate (15), Abuse (54), Neu (31)&68.68 & 65.39 \\
& Hausa (\textbf{\texttt{hau}})  & 5,311 & 818 & 2,043 & 8,172 & 3&3 & Hate (1), Abuse (31), Neu (68)&76.10 & 64.90 \\
& Igbo (\textbf{\texttt{ibo}})  & 2,872 & 442 & 1,105 & 4,419 & 3&6& Hate (8), Abuse (68), Neu (24)&80.86 & 73.46 \\
& Kinyarwanda (\textbf{\texttt{kin}}) & 3,068 & 473 & 1,180 & 4,721 & 3&3& Hate (27), Abuse (24), Neu (49)&80.00 & 78.97 \\
& Oromo (\textbf{\texttt{orm}}) & 3,201 & 493 & 1,232 & 4,926 & 4&9 & Hate (46), Abuse (13), Neu (41)&52.70 & 55.88 \\
& Somali (\textbf{\texttt{som}})  & 2,822 & 435 & 1,085 & 4,342 & 4 &7 & Hate (13), Abuse (26), Neu (59)&35.51 & 36.08 \\
& Tigrinya (\textbf{\texttt{tir}})  & 3,026 & 466 & 1,164 & 4,656 & 4&8 & Hate (60), Abuse (22), Neu (18) &45.41& 51.14 \\
& Twi (\textbf{\texttt{twi}})  & 2,534 & 390 & 975 & 3,899 & 3&3 & Hate (11), Abuse (84), Neu (5)&75.46 & 48.40 \\
& isiXhosa (\textbf{\texttt{xho}})  & 2,347 & 362 & 903 & 3,612 & 3&3 & Hate (8), Abuse (42), Neu (50)&61.21 & 54.76 \\
& Yorùbá (\textbf{\texttt{yor}})   & 2,715 & 418 & 1,045 & 4,178 & 4& 4 & Hate (5), Abuse (50), Neu (45)&76.42 & 70.79 \\
& isiZulu (\textbf{\texttt{zul}})   & 2,729 & 420 & 1,050 & 4,199 & 3&3& Hate (5), Abuse (43), Neu (52)&80.71 & 76.50 \\
\hline
\end{tabular}
}
\caption{Summary of datasets. From left to right: 1) \textbf{Task} name, 2) \textbf{Language} under each task; 3) the number of instances in \textbf{Train}, \textbf{Test}, \textbf{Dev}, and \textbf{Total}; 4) Numeber of Annotators per instance, 5) Total number of Annotators involved in the annotation, 5) \textbf{Label distribution in} \% in the dataset: Sentiment (Pos, Neg, and Neu); Emotion - (anger, disgust, fear, joy, sadness, surprise, and no emotion); and Hate speech (abuse, hate or Neutral), respectively; 6) Percentage (\%) of full agreement - all annotators agreed; and 7) overall \textbf{Cohen's Kappa} agreement.}
\label{tab:dataset}
\end{table*}

\subsection{AfriSenti - Sentiment Analysis} 
AfriSenti \cite{muhammad-etal-2023-afrisenti} is a sentiment analysis dataset targeted at 14 African languages. However, annotator-level data is available for 12 of these languages. The data was sourced from X (formerly Twitter), and the classes are either \textit{positive}, \textit{negative}, or \textit{neutral}.

\subsection{AfriEmo - Emotion Classification}
The SemEval-2025 Task 11 dataset \cite{muhammad-etal-2025-semeval} is a multilingual emotion dataset covering 32 languages from diverse domains.  
For this work, we focus on the annotator-level data available for 14 African languages. We refer to this subset as \textbf{AfriEmo}. It was annotated based on Ekman's six basic emotions \cite{ekman1999basic} (anger, disgust, fear, joy, sadness, and surprise) in a binary option with a yes or no and intensity scores. We use only the binary emotion annotation.

\subsection{AfriHate - Hate Speech Detection}
AfriHate  \cite{muhammad2025afrihate} is a hate speech dataset covering 15 African languages; 14 languages captured annotation-level data and corresponding annotator anonymous IDs. Each instance has been annotated by 3 to 4 annotators, and the classes are either \textit{abuse}, \textit{hate}, or \textit{neutral}.  The final gold label was determined by majority voting, i.e., two out of three labels or three out of four labels. In cases of a tie among four annotators (i.e., two annotators selecting one label and two selecting another), as occurs in the \texttt{amh}, \texttt{orm}, \texttt{som}, \texttt{tir}, and \texttt{yor} datasets, the 'hate' label was prioritized over other classes.

\subsection{GoEmotions and GabHate Dataset}
We evaluate two widely used English datasets: GoEmotions \cite{demszky-etal-2020-goemotions} and the Gab Hate Corpus (GabHate)  \cite{kennedy2022introducing}. GoEmotions was annotated for 27 emotion classes in a multi-label approach. However, our experiment used the six basic emotion classes, following the previous work \cite{davani_dealing_2022}. GabHate is annotated for whether the text contains hate speech or not.

\section{Experimental Setup}
This section presents the language models we fine-tuned and the evaluation setups we followed.

\subsection{Multilingual Language Models}
For the English language experiments, we use two models. In the hate speech task (GabHate), we use XLM-T \cite{barbieri-etal-2022-xlm}, an XLM-R model trained on X (Twitter) data. For the GoEmotions data, we evaluate \textit{XLM-Roberta-base} \cite{conneau2020unsupervised}. Based on benchmark results reported in the African languages dataset papers \cite{muhammad-etal-2023-afrisenti,muhammad2025afrihate,muhammad-etal-2025-semeval},  the best performance is achieved using AfroXLMR, an African-centric language model \cite{alabi-etal-2022-adapting}. AfroXLMR is an encoder-only model that extends XLM-R \cite{conneau2020unsupervised} by adding additional pretraining on 76 African languages. Our evaluation for African languages is based on AfroXLMR, as it covers all the target languages mentioned in this work. This specific version of the model can be found in Hugging Face\footnote{\url{https://huggingface.co/Davlan/afro-xlmr-large-76L}}.

\begin{table*}[h!]
\centering
\resizebox{\textwidth}{!}{
\begin{tabular}{lccccccc}
\toprule
\multirow{2}{*}{\textbf{Lang.}} & {\textbf{Baseline}}&\multicolumn{3}{c}{\textbf{Individual annotator}} & \multicolumn{3}{c}{\textbf{Clustered annotators}} \\
\cmidrule{2-2} \cmidrule(lr){3-5} \cmidrule(lr){6-8}
 & \textbf{Majority} & \textbf{Ensemble} & \textbf{Multi-label}  & \textbf{Multitask} & \textbf{Ensemble} & \textbf{Multi-label}  & \textbf{Multitask}\\
\midrule
\multicolumn{8}{l}{\textit{Sentiment analysis }}\\
ary & 50.3$_{\pm 3.0}$ & 42.2$_{\pm 1.7}$ & 29.5$_{\pm 3.9}$ & 33.0$_{\pm 2.7}$ & \textbf{53.3}$_{\pm 2.4}$ & 44.5$_{\pm 1.1}$ & 52.2$_{\pm 3.3}$  \\
pcm & 23.5$_{\pm 1.5}$ & 24.0$_{\pm 0.7}$ & 50.0$_{\pm 0.9}$ & 25.1$_{\pm 3.2}$ & 51.4$_{\pm 1.7}$ & \textbf{71.8}$_{\pm 3.5}$ & 62.5$_{\pm 2.7}$  \\
ptMZ & 26.5$_{\pm 2.1}$ & 58.9$_{\pm 0.6}$ & 26.5$_{\pm 3.3}$ & 32.9$_{\pm 3.3}$ & 56.3$_{\pm 3.9}$ & \textbf{66.3}$_{\pm 0.9}$ & 62.3$_{\pm 0.8}$  \\
yor & 51.9$_{\pm 1.1}$ & 68.2$_{\pm 0.2}$ & 70.9$_{\pm 2.2}$ & 70.5$_{\pm 2.3}$ & 65.7$_{\pm .5}$ & \textbf{72.6}$_{\pm 0.5}$ & 71.9 $_{\pm 1.6}$ \\
\midrule
\multicolumn{8}{l}{\textit{Emotion analysis}}\\
eng & 46.0$_{\pm 2.0}$ & 46.9$_{\pm 0.0}$ &55.5$_{\pm 3.9}$  & 45.2$_{\pm 2.7}$ & \textbf{60.4}$_{\pm 3.0}$ & 50.9$_{\pm 2.1}$ & 50.1$_{\pm 3.7}$  \\
hau & 69.0$_{\pm 1.9}$ & 66.3$_{\pm 0.6}$ & 67.0$_{\pm 1.7}$ & 65.0$_{\pm 0.9}$ & 66.0$_{\pm 2.1}$ & 68.0$_{\pm 0.7}$ & \textbf{71.6}$_{\pm 1.1}$  \\
orm & 41.7$_{\pm 3.2}$ & 37.0$_{\pm 1.0}$ & 61.7$_{\pm 1.9}$ & 57.77$_{\pm 0.6}$ & 46.8$_{\pm 3.3}$ & \textbf{61.1}$_{\pm 0.8}$ & 53.9$_{\pm 2.1}$  \\
som & 47.7$_{\pm 3.4}$ & 24.1$_{\pm 4.1}$ &49.6$_{\pm 0.3}$ & 47.2$_{\pm 1.5}$ & 43.0$_{\pm 1.8}$ & 45.0$_{\pm 0.9}$ & \textbf{49.8}$_{\pm 1.8}$  \\
tir & 48.0$_{\pm 2.6}$ & 36.1$_{\pm 3.1}$ & 51.9$_{\pm 2.3}$ &52.4$_{\pm 1.9}$ & 56.2$_{\pm 2.2}$ & \textbf{53.7}$_{\pm 2.7}$ & 54.4$_{\pm 1.8}$  \\
\midrule
\multicolumn{8}{l}{\textit{Hate speech analysis}}\\
amh & 71.6$_{\pm 0.8}$ & 38.6$_{\pm 2.6}$ & 71.3$_{\pm 3.6}$ & 71.5$_{\pm 2.2}$ & \textbf{73.0}$_{\pm 3.5}$ & 71.4$_{\pm 2.9}$ & 72.5$_{\pm 3.9}$  \\
eng & 47.4$_{\pm 4.2}$ & 68.5$_{\pm 2.0}$ & 68.9$_{\pm 1.9}$ & 71.9$_{\pm 2.7}$ & 69.5$_{\pm 1.5}$ & 68.6$_{\pm 1.9}$ & \textbf{73.6}$_{\pm 2.9}$  \\
ibo & 83.7$_{\pm 4.0}$ & 69.8$_{\pm 4.2}$ & 27.0$_{\pm 3.9}$ & \textbf{88.2}$_{\pm 2.3}$ & 80.4$_{\pm 3.3}$ & 87.6$_{\pm 1.7}$ & 84.0$_{\pm 0.5}$  \\
orm & 64.6$_{\pm 0.9}$ & 62.8$_{\pm 1.1}$ & 65.2$_{\pm 1.3}$ & 61.9$_{\pm 2.2}$ & 70.3$_{\pm 0.5}$ & \textbf{73.6}$_{\pm 0.3}$ & 71.9$_{\pm 0.8}$  \\
som & 50.4$_{\pm 1.1}$ & 62.7$_{\pm 1.3}$ & 58.7$_{\pm 0.7}$ & 61.9$_{\pm 0.6}$ & 64.9$_{\pm 1.8}$ & \textbf{66.4}$_{\pm 1.5}$ & 59.4$_{\pm 0.5}$  \\
tir & 67.8$_{\pm 1.5}$ & 71.4$_{\pm 1.7}$ & 70.4$_{\pm 1.5}$ & 73.3$_{\pm 0.3}$ & 64.0$_{\pm 0.1}$ & \textbf{73.9}$_{\pm 0.8}$ & 71.1$_{\pm 1.1}$  \\

\bottomrule
\end{tabular}
}
\caption{Performance comparison between \textbf{baseline}(majority vote), modeling \textbf{individual annotators}, and \textbf{clustered annotators} across aggregation methods and datasets. The results are from the number of clusters $3$ for all languages, except \textit{hau} = 5. The \textbf{boldfaced} results are the best across aggregation approaches for individual annotator modeling and after clustering.}
\label{tab:baseline}
\end{table*}

\paragraph{Training Details} The model checkpoint is accessed from the Hugging Face with the transformers library. For training, we fine-tune AfroXLMR for sequence classification using a batch size of 16, a maximum sequence length of 128, over 3 epochs, with a learning rate of 2e-5 consistently across all experiments for reproducibility. Nevertheless, our proposed approaches are not limited to a single model.

\subsection{Evaluation Setup}
We evaluate model performance using standard text classification metrics: accuracy and macro-F1, we report macro-F1 because of highly label distribution imbalance across the datasets as shown in Table \ref{tab:dataset}. Each experiment is repeated five times, and the reported results are the average across these runs. To evaluate each aggregation approach, the final prediction is obtained by applying a majority vote over the outputs of the cluster models.

\section{Experiment Results}
In this section, we present results for individual annotator modeling and the proposed clustering approach using the evaluation dataset presented in Table  \ref{tab:dataset}. 

\subsection{Individual Vs. Clustered Annotators}
The results in this section compare the languages where the clustering approach was applied.
The ensemble approach using individual annotators serves as the baseline for evaluating our proposed clustering method. We report results by comparing models trained on individual annotators with those trained on agreement-based clustered annotators. Table \ref{tab:baseline} presents the results from the \textit{individual annotator} and the \textit{clustered annotators}.

\paragraph{Modeling Individual Annotator:} While the baseline model is trained and tested on the majority vote, individual annotator modeling requires training and evaluating a separate model for each annotator using their respective labels. For example, in the sentiment analysis dataset, the \textit{pcm} language includes 11 annotators, and we train one model per annotator. The final prediction is obtained by ensembling the individual predictions through majority voting.

\paragraph{Modeling Clusters of Annotators:} Following the same approach as individual annotator modeling, we train a model for each cluster and aggregate the predictions using majority voting for final evaluation. For the English GabHate dataset, we used individual modeling results from previous work by \citet{federated-2024} and replicated the same experiment setup for clustered approaches. For the English GoEmotions dataset, we used  the ensemble of individual annotator results implemented in \citet{davani_dealing_2022}, and we train models using a cluster-based approach.

\begin{table}[h!]
\centering

\resizebox{\columnwidth}{!}{
\begin{tabular}{llccc}
\hline
\textbf{Lang.} &\textbf{\# Anno.}&\textbf{Majo.} & \textbf{Indi.}  &\textbf{ Clust.} \\
\hline
\multicolumn{5}{l}{\textit{Sentiment analysis }}\\
\texttt{ary} &9&50.3 & 42.2 & \textbf{53.3} \\
\texttt{pcm}  & 11&23.5 & 24.0& \textbf{51.4} \\
\texttt{ptMZ}  & 6&26.5 &  \textbf{58.9} &56.3\\
\texttt{yor}  &4&51.9 & \textbf{68.2} &65.7 \\
\hline
\multicolumn{5}{l}{\textit{Emotion analysis }}\\
\texttt{eng}  & 82 &46.0 & 46.9 &\textbf{60.4}\\
\texttt{hau}  & 7&69.0 & 66.3 &\textbf{66.0} \\
\texttt{orm}  & 6&41.7 & 37.0 &\textbf{46.8} \\
\texttt{som}  & 6&\textbf{47.7} & 24.1 &43.0\\
\texttt{tir}  & 6&48.0 & 36.1 &\textbf{56.2}\\
\hline
\multicolumn{5}{l}{\textit{Hate speech analysis }}\\
\texttt{amh}  & 11&71.6 & 36.6 &\textbf{73.0}\\
\texttt{eng}  & 18 &47.4 & 68.5 &\textbf{69.5}\\
\texttt{ibo}  & 6&\textbf{83.7} & 69.8 &80.4\\
\texttt{orm}  & 9&64.6 & 62.8 &\textbf{70.3}\\
\texttt{som}  & 7&50.4 & 62.7 &\textbf{64.9}\\
\texttt{tir}  & 8&67.8 & \textbf{71.4} &64.0\\
\hline
\end{tabular}
}
\caption{Zooming macro F1 score results from Table \ref{tab:baseline} for majority vote (\textbf{Majo.}), ensemble of individual annotator (\textbf{Indi.}), and ensemble of clustered annotators (\textbf{Clust.}).}
\label{tab:zooming}
\end{table}

Table \ref{tab:zooming} (summarized zoom-in results of Table \ref{tab:baseline}) shows the results from the \textit{Majority vote}, \textit{Individual Annotator} and the \textit{clustered Annotators}.

The results demonstrate that the proposed clustering approach outperforms the majority vote and the individual-annotator modeling across the datasets. Among the aggregation approaches, multi-label and multitask aggregation achieve better performance. One contributing factor to the performance gap is the uneven distribution of annotations among annotators, which results in insufficient representation during individual model training. For instance, in the \textit{pcm} sentiment analysis dataset, 11 annotators participated, but one annotator contributed 7,982 annotations (46.3\%), while another provided only 676 (3.9\%). This imbalance in annotation distribution leads to a skewed learning signal, limiting the model’s ability to effectively generalize each annotator’s perspective. As each base model in the individual ensemble is trained solely on the subset of data labeled by its corresponding annotator, annotators with fewer annotations yield weaker models, reducing the overall performance of the ensemble. In contrast, the agreement-based clustering approach mitigates this issue by forming clusters with a larger number of instances while preserving the diversity of annotator perspectives. For the same \textit{pcm} languages, the resulting clusters are \textit{cluster\_1}: 9,736, \textit{cluster\_2}: 1,266, and \textit{cluster\_3}: 8,208, from a total of 11K training instances. 

\begin{table*}[!h]
{a) \textbf{Sentiment analysis} results \par}
\par\vspace{0.4em}
\resizebox{\textwidth}{!}{\begin{tabular}{l|cccccccccccc}
\hline
Aggreg.&\texttt{amh} & \texttt{ary} &\texttt{hau}& \texttt{ibo}& \texttt{kin}&\texttt{orm}& \texttt{pcm}& \texttt{ptMZ}&\texttt{tir}&
\texttt{tso}&\texttt{twi}&\texttt{yor}\\
\hline

Maj. vote  & 53.6 & 50.3& 78.1 & 72.2 & 19.2 & \textbf{36.6} & 23.5 & 26.5 & 65.6 & 31.3 & 33.1 & 21.9\\
Ensemble  & 50.3 & \textbf{53.3}& 77.5 & 23.2 & 67.5 & 31.6 & 51.4 & 56.3 & 61.0 & 25.4 & \textbf{43.2} & 65.7 \\
Multi-label& \textbf{72.9} & 44.5& 77.1 & 37.1 & 64.7 & 25.3 & \textbf{71.8} & \textbf{66.3} & \textbf{68.4} & 30.2 & 30.6 & \textbf{72.6} \\
Multitask  & 26.7 & 52.2& \textbf{78.6} & \textbf{73.0} & \textbf{69.9} & 32.9 & 62.5 & 62.3 & 34.9 & \textbf{43.7} & 40.9 & 71.9 \\
\hline

\end{tabular}
}
\par\vspace{0.4em}
{b) \textbf{Emotion analysis} results \par}
\par\vspace{0.4em}
\resizebox{\textwidth}{!}{
\begin{tabular}{l|cccccccccccccc}
\hline
Aggreg.&\texttt{amh}&\texttt{eng} & \texttt{ary} &\texttt{hau}& \texttt{ibo}& \texttt{kin}&\texttt{orm}& \texttt{pcm}& \texttt{ptMZ}&\texttt{som}&\texttt{tir}&\texttt{xho}&\texttt{yor}&\texttt{zul}\\
\hline
Maj. vote & \textbf{67.7}&46.0 & 43.1 & 69.0 & 42.1 & 53.9 & 41.7 & \textbf{48.4} & 58.1 & 47.7  & 48.0 & 28.3 & 20.4 & 16.4\\
Ensemble & 46.2 &\textbf{60.4}& 44.2& 66.0 & 45.9 & 50.3 & 46.8 & 43.0 & 09.1 & 43.0  & 56.2 & 28.1 & 26.0 & 17.5 \\
Multi-label& 61.6&50.9 & 32.1& 68.0 & 41.8 & 53.7 & \textbf{61.1} & 48.1 & 21.7 & 45.3 & \textbf{53.7} & 30.5& \textbf{33.0} & 18.6\\
Multitask & 64.2 &50.1& \textbf{49.1}& \textbf{71.6} & \textbf{51.6} & \textbf{54.3} & 53.9 & 47.6 & \textbf{59.0} & \textbf{49.8} & 54.4 & \textbf{31.8} & 30.7 & \textbf{20.7} \\
\hline

\end{tabular}
}
\par\vspace{0.4em}
{c) \textbf{Hate speech analysis} results \par}
\par\vspace{0.4em}
\resizebox{\textwidth}{!}{\begin{tabular}{l|ccccccccccccc}
\hline
 Aggreg.&\texttt{amh}& \texttt{eng} & \texttt{ary} &\texttt{hau}& \texttt{ibo}& \texttt{kin}&\texttt{orm}& \texttt{som}&\texttt{tir}&
\texttt{twi}&\texttt{xho}&\texttt{yor}&\texttt{zul}\\
\hline
 Maj. vote & 71.7&47.4 & 69.4& 48.2 & 83.7 & 75.6 & 64.6 & 50.4 & 67.8 & 60.4 & 74.1 & 61.7 & 21.0 \\
 Ensemble & \textbf{73.0}&69.5 & 72.5& 35.9 & 80.4 & 77.8 & 70.3 & 66.9 & 73.7 & 57.4 & 78.6 & 58.5 & 82.9 \\
 Multi-label& 71.4 &68.6& 37.5& \textbf{83.2} & 87.6 & 74.6 & \textbf{73.6} & \textbf{66.4} & \textbf{73.9} & \textbf{76.7} & 76.6 & \textbf{83.3} & 83.3 \\
 Multitask & 72.5&\textbf{73.6} & \textbf{75.8}& 77.3 & \textbf{84.0} & \textbf{78.9} & 71.9 & 59.4 & 71.1 & 60.2 & \textbf{83.1} & 63.1 & \textbf{89.7} \\

\hline
\end{tabular}
}
\caption{\textbf{Sentiment}, \textbf{Emoiton}, and \textbf{Hate Speech} analysis macro F1 score results. Column names are language codes. We highlight the best aggregation approach results in bold.  
\textbf{Aggreg}.(aggregations) are the cluster aggregation approaches at the end. 
}
\label{tab:all-results}
\end{table*}

\subsection{Best Cluster Aggregation Approaches}\label{sec:best}

\paragraph{Sentiment Analysis Results}
Table~\ref{tab:all-results}(a) presents results for the Sentiment analysis dataset. For the ensemble, multi-label, and multitask models, we evaluate how well the majority vote of predicted labels from individual cluster models aligns with the pre-modeling majority vote. As a result, multi-label and multitask aggregation approaches outperform the ensemble and majority-voted labels. This suggests aggregating annotations before modeling discards valuable annotator-specific perspectives and may introduce label noise. The predicted results are more stable and accurate when clusters are modeled independently and their internal consistency is leveraged. Agreement-based clustering within a multi-label aggregation further enhances performance, achieving state-of-the-art results compared to the majority-voting benchmarks. 

Performance correlates with class distribution at the sentiment-class level: more-represented classes yield higher scores. For example, in the \textit{amh} dataset, 67\% of the instances are labeled as \textit{Negative}, achieving a macro-F1 score of 84.5\%. In contrast, the \textit{Neutral} class, which constitutes only 11\% of the data, obtains a lower score of 37.7\%. 

\paragraph{Emotion Analysis Results}
Table~\ref{tab:all-results} (b) presents results for the Emotion analysis dataset. The multitask aggregation approach performs better across most languages. This multitask out-performance is consistent with the observation that many annotators contribute only a few annotations. For instance, in the \textit{hau} dataset, each instance is annotated by 5 to 7 annotators, but less than 3\% of the data includes annotations from the 6th and 7th annotators, indicating sparsity.

At the emotion-class level, performance varies depending on label distribution. In the \textit{swa} dataset, the emotion classes \textit{disgust} and \textit{fear} account for only 7\% and 3\% of the data, respectively, making them challenging to learn. Similarly, the \textit{xho} dataset performs poorly on \textit{anger}, \textit{disgust}, and \textit{fear}, which comprise just 4\%, 0.5\%, and 2\% of the dataset, respectively. Overall, the emotion classes \textit{fear} and \textit{surprise} consistently yield the lowest performance, while \textit{joy} and \textit{sadness} achieve comparatively higher performance.

\paragraph{Hate Speech Analysis Results}
Table~\ref{tab:all-results}(c) presents results for the hate speech dataset. Consistent with prior findings in sentiment and emotion analysis, the multitask aggregation approach outperforms other methods across most languages. In contrast, the majority vote baseline consistently underperforms. Class-level performance is closely tied to label distribution: classes with more annotated examples yield better results. For instance, in the \textit{yor} dataset, the model struggles to predict the hate class, which comprises only 5\% of the data. This trend highlights the challenge of learning from imbalanced class distributions.

Finally, languages with larger training datasets generally achieve better performance. For example, \textit{hau}, which has approximately twice the data size of other languages, achieves superior results, underscoring the importance of balanced label distributions and sufficient data for effective classification.

\paragraph{Summarized Results}

Table \ref{tab:summ} shows the summarized results of the tasks and aggregation approaches at the language level presented in Table \ref{tab:all-results}. Based on the summary, the multitask aggregation approach outperforms other aggregation approaches (18/40 language datasets), with multi-label being the second best (14/40 language datasets). In addition to achieving better results with multitask and multi-label approaches, they are also efficient in terms of computational resources, such as memory and training time. This is because they train a cluster of $C$ models rather than an individual model per annotator.

\begin{table}[!h]
\centering
\begin{tabular}{p{2cm}lll}
\hline
 \textbf{Tasks}&\textbf{Models} & \textbf{Best for}&\textbf{Avg.} \\
\hline
\multirow{4}{2cm}{\textbf{Sentiment \\ analysis}} 
& Maj. vote  & 1/12 lang. & 42.7\\
& Ensemble   & 2/12 lang. & 50.5\\
& \cellcolor[gray]{0.9}Multi-label& \cellcolor[gray]{0.9}5/12 lang.  &  \cellcolor[gray]{0.9}55.1\\
& Multitask  & 4/12 lang.   & 54.1 \\
\hline
\multirow{4}{2cm}{\textbf{Emotion \\ analysis}} 
& Maj. vote  & 2/14 lang.  & 45.1 \\
& Ensemble   & 1/14 lang.   & 41.6\\
& Multi-label & 3/14 lang.  & 45.7 \\
& \cellcolor[gray]{0.9}Multitask  & \cellcolor[gray]{0.9}8/14 lang.  & \cellcolor[gray]{0.9}\textbf{49.2} \\
\hline
\multirow{4}{2cm}{\textbf{HateSpeech \\ analysis}} 
& Maj. vote  & 0/13 lang.  & 61.2 \\
& Ensemble   & 1/13 lang.  & 69.0 \\
& Multi-label& 6/13 lang.  & 73.6 \\
& \cellcolor[gray]{0.9}Multitask  & \cellcolor[gray]{0.9}6/13 lang.  & \cellcolor[gray]{0.9}\textbf{73.9} \\
\hline
\end{tabular}
\caption{Summarized results across modeling disagreement approaches and datasets. The \textbf{Best for} column indicates the number of languages (out of the total evaluated) for which a specific model type achieved the best performance.  \textbf{Avg.} is average macro F1 score across total number languages. }
\label{tab:summ}
\end{table}

\section{Discussion and Analysis}

\paragraph{Individual vs. Agreement-based Clustering:}
Our agreement-based clustering approach is more robust than individual annotator modeling and majority vote baselines. By clustering annotators based on their annotation behavior, we capture individual perspectives that improve performance, provide stable training, and reduce computational resources. This advantage arises because some annotators label very few instances, while others label nearly all, making individual annotator-level modeling inefficient and inconsistent.

As expected, individual annotator modeling is the most resource-intensive approach, requiring training $N$ separate models (where $N$ is the number of annotators), each with varying training sizes. In contrast, the clustering approach requires training fewer models (e.g., 3 or 5 in our experiments), resulting in faster and more efficient training. Notably, the multi-label and multitask model training and aggregation approaches offer a favorable trade-off, delivering strong performance without significantly increasing training time.  

\paragraph{Performance in Terms of Agreement:} 
We investigate how inter-annotator agreement impacts the performance of different aggregation approaches. Specifically, we examine whether there is a correlation between model performance and overall annotator agreement. Our findings indicate a general trend: higher inter-annotator agreement corresponds to better model performance. For instance, languages such as \textit{hau}, \textit{tir}, and \textit{yor}, which exhibit relatively high Cohen's Kappa scores of 61.5\%, 61.8\%, and 63.0\%, respectively, achieve superior results in the sentiment analysis task. This suggests a direct relationship between annotation consistency and model effectiveness. However, we note that these comparisons are not direct due to variation in data size across languages, which may also influence performance outcomes.

\paragraph{Number of Annotators vs. Agreement:} 
We examine whether the number of annotators involved in the annotation process is related to overall pairwise agreement, as measured by Cohen's Kappa. We observe a trend in which an increase in the number of annotators tends to correspond to a decrease in agreement scores. For instance, in the sentiment analysis dataset, \textit{ptMZ}, \textit{ary}, and \textit{pcm} involved 6, 9, and 11 annotators, respectively, but achieved relatively lower Cohen’s Kappa scores of 33.8\%, 52.6\%, and 46.4\%. Similarly, in the hate speech dataset, languages such as \textit{som} (7 annotators), \textit{tir} (8), \textit{orm} (9), and \textit{amh} (11) recorded lower agreement scores of 36.1\%, 51.1\%, 55.9\%, and 59.7\%, respectively, lower than other languages with fewer annotators. In terms of performance, datasets with four annotators and a majority vote threshold requiring at least two votes tend to yield better results. This is evident in the AfriHate dataset, where languages such as \textit{amh}, \textit{tir}, \textit{orm}, and \textit{som} exhibit relatively stronger model performance under this setting.

\paragraph{Performance Across Tasks:} 
As presented in Table~\ref{tab:summ}, a comparison across the three tasks reveals that the emotion classification task consistently performs worse than sentiment analysis and hate speech detection tasks. Sentiment task records an average macro F1 score of 55.1\% across 12 languages, whereas Emotion and hate speech analysis tasks achieve 49.2\% and 73.9\%, respectively. This performance gap can be primarily attributed to the inherent complexity of the multi-label emotion classification task. 

\paragraph{Computational Resources}
Individual annotator modeling is computationally very demanding, particularly in terms of memory requirements and training time, since a separate model must be trained for each annotator. For example, the model we used for African languages requires 2.24 GB of memory. Training a separate model for each of the 11 annotators in the \texttt{pcm} languages would therefore require 35 minutes (depending on the number of instances per annotator) on a single GPU  and approximately 2.24GB×11 of memory size. The same scaling applies to crowd-sourced datasets such as the English GoEmotions dataset, which involves 82 annotators. To improve efficiency and performance, our proposed clustering of annotators based on their agreement is effective.

\section{Conclusion}
In this work, we presented an agreement-based annotator clustering approach for modeling annotation disagreement in subjective NLP tasks, providing a more effective alternative to aggregating annotations through majority voting.
The multitask and multi-label aggregation approaches outperform both the majority-vote and individual-annotator modeling approaches. In addition to achieving better performance, these approaches are also more computationally efficient, as they require training only a cluster of $C$ models rather than one model per annotator.
The work addressed key limitations highlighted in previous works, which involved training separate models for each annotator. In contrast, our approach efficiently captured diverse annotator perspectives while reducing training overhead and deferring label aggregation until the final prediction stage. This makes it especially useful for real-world applications such as social media monitoring, opinion mining, and others, where understanding subjectivity and disagreement is crucial. Finally, our results across diverse datasets demonstrate that leveraging the full spectrum of annotator input, rather than collapsing annotations into a single majority-voted label, significantly enhances classification performance in subjective NLP tasks while preserving individual annotator perspectives.

\section*{Limitations}
Our work is not without limitations.

\paragraph{Model Varieties:}
Due to high computational resources across languages, tasks, and the number of annotators, our experimental setup is limited to a single model evaluation with the least hyperparameter settings. Each downstream NLP task and aggregation approach might need different hyperparameter settings. In the future, this work can be extended to evaluate multiple variants of pre-trained language models and both open-source and proprietary LLMs.

\paragraph{Number of Clusters:} Our agreement-based clustering approach groups annotators into $C$ clusters, where the number of clusters $C$ could be determined based on the availability of computational resources and data factors. Our clustering principle works for any number of clusters and can be extended to any custom number of clusters based on the computational resource availability. This is because, as the number of clusters increases, computational resources are also increased. However, we recommend using an odd number of clusters and applying the majority vote rule to the final label for evaluation after modeling each cluster separately. The clustering principle works for any number of clusters and can be extended to a more custom number of clusters based on the training resource availability. For example, if the dataset is annotated by a fixed number of annotators, equally, a different number of clusters may work better. However, in our work, we only used clustering into 3 and 5 clusters. 

In a dataset annotated by many annotators, where each instance (text) is annotated equally and requires a custom number of clusters, this method is not applied because the number of annotations for a text equals the number of clusters. Exploring Other clustering approaches, such as Hierarchical and Spectral clustering of annotators, clustering algorithms like Krippendorff’s alpha, and clustering methods for annotators that do not have common annotated instances, is an area of open research that will also be an open area for further investigation. 

\paragraph{Aggregation Methods:} This work, as well as previously conducted works that are mentioned in the related works section, uses a majority vote to decide the final predicted label after modeling the perspectives of annotators. In individual modeling, we train a model for each annotator separately and evaluate using test sets. It does not matter how many annotators annotate each instance, as each annotator uses their own annotation to train an individual model. During clustering, no further individual annotation is provided; annotators are grouped into a cluster, and only cluster-level annotation is provided. We can assume one cluster as an individual annotator. However, other aggregation techniques should be explored, such as using annotator soft labels as gold labels for the final evaluation.

\section*{Acknowledgments}
We are very grateful to the editors and anonymous reviewers for their constructive comments. The work was done with partial support from the Mexican Government through the grant A1-S-47854 of CONACYT, Mexico, grants 20241816, 20241819,  and 20240951 of the Secretaría de Investigación y Posgrado of the Instituto Politécnico Nacional, Mexico. The authors thank the CONACYT for the computing resources brought to them through the Plataforma de Aprendizaje Profundo para Tecnologías del Lenguaje of the Laboratorio de Supercómputo of the INAOE, Mexico and acknowledge the support of Microsoft through the Microsoft Latin America PhD Award. We thank the authors of the dataset used in this paper, who made these subjective NLP task datasets, including individual annotator levels, available. We appreciate and recommend releasing such individual annotator levels for the subjectivity study.
\bibliography{tacl2021}

@InProceedings{978-3-032-12993-2_29,
author="Parmar, Unnati
and Modh, Jatin",
editor="Iglesias, Andres
and Shin, Jungpil
and Bhatt, Nityesh
and Joshi, Amit",
title={{Advancements in Sentiment Analysis: A Comprehensive Survey of Techniques, Models, and Real-World Applications}},
booktitle="Information Systems for Intelligent Systems",
year="2026",
publisher="Springer Nature Switzerland",
address="Cham",
pages="301--308",
isbn="978-3-032-12993-2",
doi="https://doi.org/10.1007/978-3-032-12993-2_29"
}

@article{afriyie2026sentiment,
  title={Sentiment analysis based on deep learning approaches for text classification},
  author={Afriyie, Yaw and Weyori, Benjamin A},
  journal={Journal of Electrical Systems and Information Technology},
  volume={13},
  number={1},
  pages={14},
  year={2026},
  DOI="https://link.springer.com/article/10.1186/s43067-026-00322-4",
  publisher={Springer}
}

@inproceedings{beck-etal-2024-sensitivity,
    title = {{Sensitivity, Performance, Robustness: Deconstructing the Effect of Sociodemographic Prompting}},
    author = "Beck, Tilman  and
      Schuff, Hendrik  and
      Lauscher, Anne  and
      Gurevych, Iryna",
    editor = "Graham, Yvette  and
      Purver, Matthew",
    booktitle = "Proceedings of the 18th Conference of the European Chapter of the Association for Computational Linguistics (Volume 1: Long Papers)",
    month = mar,
    year = "2024",
    address = "St. Julian{'}s, Malta",
    publisher = "Association for Computational Linguistics",
    url = "https://aclanthology.org/2024.eacl-long.159/",
    doi = "10.18653/v1/2024.eacl-long.159",
    pages = "2589--2615"
}

@inproceedings{kapil-ekbal-2024-survey,
    title = {{A Survey on Combating Hate Speech through Detection and Prevention in {E}nglish}},
    author = "Kapil, Prashant  and
      Ekbal, Asif",
    editor = "Lalitha Devi, Sobha  and
      Arora, Karunesh",
    booktitle = "Proceedings of the 21st International Conference on Natural Language Processing (ICON)",
    month = dec,
    year = "2024",
    address = "AU-KBC Research Centre, Chennai, India",
    publisher = "NLP Association of India (NLPAI)",
    url = "https://aclanthology.org/2024.icon-1.57/",
    pages = "485--501"
}

@inproceedings{Himabindu-2015,
author = {Himabindu Lakkaraju and Jure Leskovec and Jon Kleinberg and Sendhil Mullainathan},
title = {A Bayesian Framework for Modeling Human Evaluations},
booktitle = {Proceedings of the 2015 SIAM International Conference on Data Mining (SDM)},
publisher ={Society for Industrial and Applied Mathematics},
year = {2015},
pages = {181-189},
doi = {10.1137/1.9781611974010.21},
URL = {https://epubs.siam.org/doi/abs/10.1137/1.9781611974010.21}
}

@inproceedings{Matteo-community-based-bayesian,
author = {Venanzi, Matteo and Guiver, John and Kazai, Gabriella and Kohli, Pushmeet and Shokouhi, Milad},
title = {Community-based bayesian aggregation models for crowdsourcing},
year = {2014},
isbn = {9781450327442},
publisher = {Association for Computing Machinery},
address = {New York, NY, USA},
url = {https://doi.org/10.1145/2566486.2567989},
doi = {10.1145/2566486.2567989},
booktitle = {Proceedings of the 23rd International Conference on World Wide Web},
pages = {155–164},
numpages = {10},
location = {Seoul, Korea},
series = {WWW '14}
}

@inproceedings{jury-learning-gordon,
author = {Gordon, Mitchell L. and Lam, Michelle S. and Park, Joon Sung and Patel, Kayur and Hancock, Jeff and Hashimoto, Tatsunori and Bernstein, Michael S.},
title = {Jury Learning: Integrating Dissenting Voices into Machine Learning Models},
year = {2022},
isbn = {9781450391573},
publisher = {Association for Computing Machinery},
address = {New York, NY, USA},
url = {https://doi.org/10.1145/3491102.3502004},
doi = {10.1145/3491102.3502004},
booktitle = {Proceedings of the 2022 CHI Conference on Human Factors in Computing Systems},
articleno = {115},
numpages = {19},
location = {New Orleans, LA, USA},
series = {CHI '22}
}

@inproceedings{weerasooriya-etal-2022-improving,
    title = "Improving Label Quality by Jointly Modeling Items and Annotators",
    author = "Weerasooriya, Tharindu Cyril  and
      Ororbia, Alexander  and
      Homan, Christopher",
    editor = "Abercrombie, Gavin  and
      Basile, Valerio  and
      Tonelli, Sara  and
      Rieser, Verena  and
      Uma, Alexandra",
    booktitle = "Proceedings of the 1st Workshop on Perspectivist Approaches to NLP @LREC2022",
    month = jun,
    year = "2022",
    address = "Marseille, France",
    publisher = "European Language Resources Association",
    url = "https://aclanthology.org/2022.nlperspectives-1.12/",
    pages = "95--99"
}

@article{kennedy2022introducing,
  title={Introducing the Gab Hate Corpus: defining and applying hate-based rhetoric to social media posts at scale},
  author={Kennedy, Brendan and Atari, Mohammad and Davani, Aida Mostafazadeh and Yeh, Leigh and Omrani, Ali and Kim, Yehsong and Coombs, Kris and Havaldar, Shreya and Portillo-Wightman, Gwenyth and Gonzalez, Elaine and others},
  journal={Language Resources and Evaluation},
  pages={1--30},
  year={2022},
url={https://osf.io/preprints/psyarxiv/hqjxn_v1},
  publisher={Springer}
}

@article{federated-2024,
    title = "Federated Learning for Exploiting Annotators{'} Disagreements in Natural Language Processing",
    author = "Rodr{\'\i}guez-Barroso, Nuria  and
      C{\'a}mara, Eugenio Mart{\'\i}nez  and
      Collados, Jose Camacho  and
      Luz{\'o}n, M. Victoria  and
      Herrera, Francisco",
    journal = "Transactions of the Association for Computational Linguistics",
    volume = "12",
    year = "2024",
    address = "Cambridge, MA",
    publisher = "MIT Press",
    url = "https://aclanthology.org/2024.tacl-1.35",
    doi = "10.1162/tacl_a_00664",
    pages = "630--648"
}

@Article{hayat2022modeling,
AUTHOR = {Hayat, Hassan and Ventura, Carles and Lapedriza, Agata},
TITLE = {Modeling Subjective Affect Annotations with Multi-Task Learning},
JOURNAL = {Sensors},
VOLUME = {22},
YEAR = {2022},
NUMBER = {14},
ARTICLE-NUMBER = {5245},
URL = {https://www.mdpi.com/1424-8220/22/14/5245},
PubMedID = {35890925},
ISSN = {1424-8220},
DOI = {10.3390/s22145245}
}

@inproceedings{frenda-etal-2023-epic,
    title = "{EPIC}: Multi-Perspective Annotation of a Corpus of Irony",
    author = "Frenda, Simona  and Pedrani, Alessandro  and    Basile, Valerio  and
      Lo, Soda Marem  and    Cignarella, Alessandra Teresa  and  Panizzon, Raffaella  and Marco, Cristina  and Scarlini, Bianca  and  Patti, Viviana  and Bosco, Cristina  and  Bernardi, Davide",
    booktitle = "Proceedings of the 61st Annual Meeting of the Association for Computational Linguistics (Volume 1: Long Papers)",
    month = jul,
    year = "2023",
    address = "Toronto, Canada",
    publisher = "Association for Computational Linguistics",
    url = "https://aclanthology.org/2023.acl-long.774/",
    doi = "10.18653/v1/2023.acl-long.774",
    pages = "13844--13857"
}

@inproceedings{akhtar2020modeling,
  title={Modeling annotator perspective and polarized opinions to improve hate speech detection},
  author={Akhtar, Sohail and Basile, Valerio and Patti, Viviana},
  booktitle={Proceedings of the AAAI conference on human computation and crowdsourcing},
  volume={8},
  pages={151--154},
doi={https://doi.org/10.1609/hcomp.v8i1.7473},
  year={2020}
}

@inproceedings{lo2023hierarchical,
  title={Hierarchical Clustering of Label-based Annotator Representations for Mining Perspectives.},
  author={Lo, Soda Marem and Basile, Valerio},
  booktitle={NLPerspectives@ ECAI},
url={https://ceur-ws.org/Vol-3494/paper8.pdf},
  year={2023}
}

@inproceedings{parting-crowds-2016,
author = {Kairam, Sanjay and Heer, Jeffrey},
title = {Parting Crowds: Characterizing Divergent Interpretations in Crowdsourced Annotation Tasks},
booktitle ={Proceedings of the 19th ACM Conference on Computer-Supported Cooperative Work and Social Computing
},
year = {2016},
isbn = {9781450335928},
publisher = {Association for Computing Machinery},
address = {New York, NY, USA},
url = {https://doi.org/10.1145/2818048.2820016},
doi = {10.1145/2818048.2820016},
pages = {1637–1648},
numpages = {12},
location = {San Francisco, California, USA},
series = {CSCW '16}
}

@inproceedings{liu-etal-2019-multi-lingual,
    title = "Multi-lingual {W}ikipedia Summarization and Title Generation On Low Resource Corpus",
    author = "Liu, Wei  and Li, Lei  and  Huang, Zuying  and Liu, Yinan",
    booktitle = "Proceedings of the Workshop MultiLing 2019: Summarization Across Languages, Genres and Sources",
    month = sep,
    year = "2019",
    address = "Varna, Bulgaria",
    publisher = "INCOMA Ltd.",
    url = "https://aclanthology.org/W19-8904/",
    doi = "10.26615/978-954-452-058-8_004",
    pages = "17--25"
}

@inproceedings{simpson-etal-2019-predicting,
    title = "Predicting Humorousness and Metaphor Novelty with {G}aussian Process Preference Learning",
    author = "Simpson, Edwin  and    Do Dinh, Erik-L{\^a}n  and Miller, Tristan  and   Gurevych, Iryna",
    booktitle = "Proceedings of the 57th Annual Meeting of the Association for Computational Linguistics",
    month = jul,
    year = "2019",
    address = "Florence, Italy",
    publisher = "Association for Computational Linguistics",
    url = "https://aclanthology.org/P19-1572/",
    doi = "10.18653/v1/P19-1572",
    pages = "5716--5728"
}

@article{zhang2022examining,
title = {Examining and mitigating gender bias in text emotion detection task},
journal = {Neurocomputing},
author = { Odbal and Guanhong Zhang and Sophia Ananiadou},
volume = {493},
pages = {422-434},
year = {2022},
issn = {0925-2312},
doi = {https://doi.org/10.1016/j.neucom.2022.04.057},
url = {https://www.sciencedirect.com/science/article/pii/S0925231222004374},
}

@inproceedings{demszky-etal-2020-goemotions,
    title = "{G}o{E}motions: A Dataset of Fine-Grained Emotions",
    author = "Demszky, Dorottya  and      Movshovitz-Attias, Dana  and  Ko, Jeongwoo  and Cowen, Alan  and
      Nemade, Gaurav  and     Ravi, Sujith",
    booktitle = "Proceedings of the 58th Annual Meeting of the Association for Computational Linguistics",
    month = jul,
    year = "2020",
    address = "Online",
    publisher = "Association for Computational Linguistics",
    url = "https://aclanthology.org/2020.acl-main.372/",
    doi = "10.18653/v1/2020.acl-main.372",
    pages = "4040--4054"
}

@inproceedings{daval-frerot-weis-2020-wmd,
    title = "{WMD} at {S}em{E}val-2020 Tasks 7 and 11: Assessing Humor and Propaganda Using Unsupervised Data Augmentation",
    author = "Daval-Frerot, Guillaume  and      Weis, Yannick",
    booktitle = "Proceedings of the Fourteenth Workshop on Semantic Evaluation",
    month = dec,
    year = "2020",
    address = "Barcelona (online)",
    publisher = "International Committee for Computational Linguistics",
    url = "https://aclanthology.org/2020.semeval-1.246/",
    doi = "10.18653/v1/2020.semeval-1.246",
    pages = "1865--1874"
}

@article{martinez2018overview,
  title={Overview of TASS 2018: Opinions, health and emotions},
  author={Mart{\'\i}nez C{\'a}mara, Eugenio and Almeida-Cruz, Yudivi{\'a}n and D{\'\i}az Galiano, Manuel Carlos and Est{\'e}vez-Velarde, Suilan and Garc{\'\i}a Cumbreras, Miguel {\'A}ngel and Garc{\'\i}a Vega, Manuel and Guti{\'e}rrez, Yoan and Montejo R{\'a}ez, Arturo and Montoyo, Andr{\'e}s and Munoz, Rafael and others},
  year={2018},
  publisher={Sun SITE Central Europe},
url={http://tass.sepln.org/2018/}
}

@inproceedings{sandri-etal-2023-dont,
    title = "Why Don`t You Do It Right? Analysing Annotators' Disagreement in Subjective Tasks",
    author = "Sandri, Marta  and
      Leonardelli, Elisa  and
      Tonelli, Sara  and
      Jezek, Elisabetta",
    editor = "Vlachos, Andreas  and
      Augenstein, Isabelle",
    booktitle = "Proceedings of the 17th Conference of the European Chapter of the Association for Computational Linguistics",
    month = may,
    year = "2023",
    address = "Dubrovnik, Croatia",
    publisher = "Association for Computational Linguistics",
    url = "https://aclanthology.org/2023.eacl-main.178/",
    doi = "10.18653/v1/2023.eacl-main.178",
    pages = "2428--2441"
}

@inproceedings{Wan-Kim-Kang-2023, 
author = {Wan, Ruyuan and Kim, Jaehyung and Kang, Dongyeop},
title = {Everyone's voice matters: quantifying annotation disagreement using demographic information},
year = {2023},
isbn = {978-1-57735-880-0},
publisher = {AAAI Press},
url = {https://doi.org/10.1609/aaai.v37i12.26698},
doi = {10.1609/aaai.v37i12.26698},
booktitle = {Proceedings of the Thirty-Seventh AAAI Conference on Artificial Intelligence}
}

@inproceedings{mokhberian-etal-2024-capturing,
    title = "Capturing Perspectives of Crowdsourced Annotators in Subjective Learning Tasks",
    author = "Mokhberian, Negar  and Marmarelis, Myrl  and
      Hopp, Frederic  and Basile, Valerio  and Morstatter, Fred  and
      Lerman, Kristina",
    booktitle = "Proceedings of the 2024 Conference of the North American Chapter of the Association for Computational Linguistics: Human Language Technologies (Volume 1: Long Papers)",
    month = jun,
    year = "2024",
    address = "Mexico City, Mexico",
    publisher = "Association for Computational Linguistics",
    url = "https://aclanthology.org/2024.naacl-long.407",
    doi = "10.18653/v1/2024.naacl-long.407",
    pages = "7337--7349"
}

@article{denton2021whose,
  title={Whose ground truth? accounting for individual and collective identities underlying dataset annotation},
  author={Denton, Remi and D{\'\i}az, Mark and Kivlichan, Ian and Prabhakaran, Vinodkumar and Rosen, Rachel},
  journal={arXiv preprint arXiv:2112.04554},
url={https://arxiv.org/abs/2112.04554},
  year={2021}
}

@inproceedings{xu-etal-2024-leveraging,
    title = "Leveraging Annotator Disagreement for Text Classification",
    author = {Xu, Jin  and
      Theune, Mari{\"e}t  and
      Braun, Daniel},
    booktitle = "Proceedings of the 7th International Conference on Natural Language and Speech Processing (ICNLSP 2024)",
    month = oct,
    year = "2024",
    address = "Trento",
    publisher = "Association for Computational Linguistics",
    url = "https://aclanthology.org/2024.icnlsp-1.1",
    pages = "1--10",
}

@inproceedings{lee-etal-2024-exploring-cross,
    title = "Exploring Cross-Cultural Differences in {E}nglish Hate Speech Annotations: From Dataset Construction to Analysis",
    author = "Lee, Nayeon  and Jung, Chani  and Myung, Junho  and Jin, Jiho  and  Camacho-Collados, Jose  and
      Kim, Juho  and  Oh, Alice",
    booktitle = "Proceedings of the 2024 Conference of the North American Chapter of the Association for Computational Linguistics: Human Language Technologies (Volume 1: Long Papers)",
    month = jun,
    year = "2024",
    address = "Mexico City, Mexico",
    publisher = "Association for Computational Linguistics",
    url = "https://aclanthology.org/2024.naacl-long.236/",
    doi = "10.18653/v1/2024.naacl-long.236",
    pages = "4205--4224"
}

@article{Uma-survey-2022,
author = {Uma, Alexandra N. and Fornaciari, Tommaso and Hovy, Dirk and Paun, Silviu and Plank, Barbara and Poesio, Massimo},
title = {Learning from Disagreement: A Survey},
year = {2022},
issue_date = {Jan 2022},
publisher = {AI Access Foundation},
address = {El Segundo, CA, USA},
volume = {72},
issn = {1076-9757},
url = {https://doi.org/10.1613/jair.1.12752},
doi = {10.1613/jair.1.12752},
journal = {J. Artif. Int. Res.},
month = jan,
pages = {1385–1470},
numpages = {86}
}

@misc{Jigsaw-2019,
    author = {Cjadams and Daniel Borkan and inversion and Jeffrey Sorensen and Lucas Dixon and Lucy Vasserman and nithum},
    title = {Jigsaw Unintended Bias in Toxicity Classification},
    year = {2019},
    url = {https://kaggle.com/competitions/jigsaw-unintended-bias-in-toxicity-classification},
    note = {Kaggle}
}

@inproceedings{diaz2020biases,
author = {Diaz, Mark and Johnson, Isaac and Lazar, Amanda and Piper, Anne Marie and Gergle, Darren},
title = {Addressing Age-Related Bias in Sentiment Analysis},
year = {2018},
isbn = {9781450356206},
publisher = {Association for Computing Machinery},
url = {https://doi.org/10.1145/3173574.3173986},
doi = {10.1145/3173574.3173986},
booktitle = {Proceedings of the 2018 CHI Conference on Human Factors in Computing Systems},
pages = {1–14},
numpages = {14},
location = {Montreal QC, Canada}
}

@misc{hatexplain,
      title={HateXplain: A Benchmark Dataset for Explainable Hate Speech Detection}, 
      author={Binny Mathew and Punyajoy Saha and Seid Muhie Yimam and Chris Biemann and Pawan Goyal and Animesh Mukherjee},
      year={2022},
      eprint={2012.10289},
      archivePrefix={arXiv},
      primaryClass={cs.CL},
      url={https://arxiv.org/abs/2012.10289}, 
}

@article{ekman1999basic,
  title={Basic emotions},
  author={Ekman, Paul and others},
  journal={Handbook of cognition and emotion},
  volume={98},
  number={45-60},
  pages={16},
  doi={10.1002/0470013494},
  year={1999}
}

@inproceedings{muhammad-etal-2023-afrisenti,
    title = "{A}fri{S}enti: A {T}witter Sentiment Analysis Benchmark for {A}frican Languages",
    author = "Muhammad, Shamsuddeen Hassan  and
      Abdulmumin, Idris  and
      Ayele, Abinew Ali  and
      Ousidhoum, Nedjma  and
      Adelani, David Ifeoluwa  and
      Yimam, Seid Muhie  and
      Ahmad, Ibrahim Sa'id  and
      Beloucif, Meriem  and
      Mohammad, Saif M.  and
      Ruder, Sebastian  and
      Hourrane, Oumaima  and
      Brazdil, Pavel  and
      Jorge, Alipio  and
      Ali, Felermino D{\'a}rio M{\'a}rio Ant{\'o}nio  and
      David, Davis  and
      Osei, Salomey  and
      Shehu Bello, Bello  and
      Ibrahim, Falalu  and
      Gwadabe, Tajuddeen  and
      Rutunda, Samuel  and
      Belay, Tadesse  and
      Messelle, Wendimu Baye  and
      Balcha, Hailu Beshada  and
      Chala, Sisay Adugna  and
      Gebremichael, Hagos Tesfahun  and
      Opoku, Bernard  and
      Arthur, Stephen",
    booktitle = "Proceedings of the 2023 Conference on Empirical Methods in Natural Language Processing",
    month = dec,
    year = "2023",
    address = "Singapore",
    publisher = "Association for Computational Linguistics",
    url = "https://aclanthology.org/2023.emnlp-main.862/",
    doi = "10.18653/v1/2023.emnlp-main.862",
    pages = "13968--13981"
}

@inproceedings{rottger-etal-2022-two,
    title = "Two Contrasting Data Annotation Paradigms for Subjective {NLP} Tasks",
    author = {R{\"o}ttger, Paul  and
      Vidgen, Bertie  and
      Hovy, Dirk  and
      Pierrehumbert, Janet},
    booktitle = "Proceedings of the 2022 Conference of the North American Chapter of the Association for Computational Linguistics: Human Language Technologies",
    month = jul,
    year = "2022",
    address = "Seattle, United States",
    publisher = "Association for Computational Linguistics",
    url = "https://aclanthology.org/2022.naacl-main.13/",
    doi = "10.18653/v1/2022.naacl-main.13",
    pages = "175--190"
}

@inproceedings{muhammad-etal-2025-semeval,
title = "{S}em{E}val-2025 Task 11: Bridging the Gap in Text-Based Emotion Detection",
author = "Muhammad, Shamsuddeen Hassan and Ousidhoum, Nedjma and Abdulmumin, Idris and Yimam, Seid Muhie and Wahle, Jan Philip and Ruas, Terry and Beloucif, Meriem and De Kock, Christine and Belay, Tadesse Destaw and Ahmad, Ibrahim Said and Surange, Nirmal and Teodorescu, Daniela and Adelani, David Ifeoluwa and Aji, Alham Fikri and Ali, Felermino and Araujo, Vladimir and Ayele, Abinew Ali and Ignat, Oana and Panchenko, Alexander and Zhou, Yi and Mohammad, Saif M.",
booktitle = "Proceedings of the 19th International Workshop on Semantic Evaluation (SemEval-2025)",
month = july,
year = "2025",
address = "Vienna, Austria",
publisher = "Association for Computational Linguistics",
url = "https://arxiv.org/abs/2503.07269",
doi = "",
pages = ""
}

@inproceedings{muhammad2025afrihate,
    title = "{A}fri{H}ate: A Multilingual Collection of Hate Speech and Abusive Language Datasets for {A}frican Languages",
    author = {Muhammad, Shamsuddeen Hassan  and Abdulmumin, Idris  and Ayele, Abinew Ali  and Adelani, David Ifeoluwa  and
      Ahmad, Ibrahim Said  and Aliyu, Saminu Mohammad  and R{\"o}ttger, Paul  and Oppong, Abigail  and Bukula, Andiswa  and
      Chukwuneke, Chiamaka Ijeoma  and
      Jibril, Ebrahim Chekol  and
      Ismail, Elyas Abdi  and
      Alemneh, Esubalew  and
      Gebremichael, Hagos Tesfahun  and
      Aliyu, Lukman Jibril  and
      Beloucif, Meriem  and
      Hourrane, Oumaima  and
      Mabuya, Rooweither  and
      Osei, Salomey  and
      Rutunda, Samuel  and
      Belay, Tadesse Destaw  and
      Guge, Tadesse Kebede  and
      Asfaw, Tesfa Tegegne  and
      Wanzare, Lilian Diana Awuor  and
      Onyango, Nelson Odhiambo  and
      Yimam, Seid Muhie  and
      Ousidhoum, Nedjma},
    booktitle = "Proceedings of the 2025 Conference of the Nations of the Americas Chapter of the Association for Computational Linguistics: Human Language Technologies (Volume 1: Long Papers)",
    month = apr,
    year = "2025",
    address = "Albuquerque, New Mexico",
    publisher = "Association for Computational Linguistics",
    url = "https://aclanthology.org/2025.naacl-long.92/",
    doi = "10.18653/v1/2025.naacl-long.92",
    pages = "1854--1871",
    ISBN = "979-8-89176-189-6"
}

@proceedings{comedi-ws-2025-1,
    title = "Proceedings of Context and Meaning: Navigating Disagreements in NLP Annotation",
    editor = "Roth, Michael  and
      Schlechtweg, Dominik",
    month = jan,
    year = "2025",
    address = "Abu Dhabi, UAE",
    publisher = "International Committee on Computational Linguistics",
    url = "https://aclanthology.org/2025.comedi-1.0/"
}

@misc{khurana2024crowdcalibrator,
      title={Crowd-Calibrator: Can Annotator Disagreement Inform Calibration in Subjective Tasks?}, 
      author={Urja Khurana and Eric Nalisnick and Antske Fokkens and Swabha Swayamdipta},
      year={2024},
      eprint={2408.14141},
      archivePrefix={arXiv},
      primaryClass={cs.CL},
      url={https://arxiv.org/abs/2408.14141}, 
}

@inproceedings{wang-plank-2023-actor,
    title = "{ACTOR}: Active Learning with Annotator-specific Classification Heads to Embrace Human Label Variation",
    author = "Wang, Xinpeng  and      Plank, Barbara",
    booktitle = "Proceedings of the 2023 Conference on Empirical Methods in Natural Language Processing",
    month = dec,
    year = "2023",
    address = "Singapore",
    publisher = "Association for Computational Linguistics",
    url = "https://aclanthology.org/2023.emnlp-main.126/",
    doi = "10.18653/v1/2023.emnlp-main.126",
    pages = "2046--2052"
}

@inproceedings{van-aken-etal-2018-challenges,
    title = "Challenges for Toxic Comment Classification: An In-Depth Error Analysis",
    author = {van Aken, Betty  and  Risch, Julian  and Krestel, Ralf  and
      L{\"o}ser, Alexander},
    booktitle = "Proceedings of the 2nd Workshop on Abusive Language Online ({ALW}2)",
    month = oct,
    year = "2018",
    address = "Brussels, Belgium",
    publisher = "Association for Computational Linguistics",
    url = "https://aclanthology.org/W18-5105/",
    doi = "10.18653/v1/W18-5105",
    pages = "33--42"
}

@inproceedings{rizzi-etal-2025-bunch,
    title = "Is a bunch of words enough to detect disagreement in hateful content?",
    author = "Rizzi, Giulia  and
      Rosso, Paolo  and
      Fersini, Elisabetta",
    editor = "Roth, Michael  and
      Schlechtweg, Dominik",
    booktitle = "Proceedings of Context and Meaning: Navigating Disagreements in NLP Annotation",
    month = jan,
    year = "2025",
    address = "Abu Dhabi, UAE",
    publisher = "International Committee on Computational Linguistics",
    url = "https://aclanthology.org/2025.comedi-1.1/",
    pages = "1--11"
}

@inproceedings{leonardelli-etal-2023-semeval,
    title = "{S}em{E}val-2023 Task 11: Learning with Disagreements ({L}e{W}i{D}i)",
    author = "Leonardelli, Elisa  and       Abercrombie, Gavin  and
      Almanea, Dina  and      Basile, Valerio  and       Fornaciari, Tommaso  and
      Plank, Barbara  and      Rieser, Verena  and
      Uma, Alexandra  and      Poesio, Massimo",
    booktitle = "Proceedings of the 17th International Workshop on Semantic Evaluation (SemEval-2023)",
    month = jul,
    year = "2023",
    address = "Toronto, Canada",
    publisher = "Association for Computational Linguistics",
    url = "https://aclanthology.org/2023.semeval-1.314/",
    doi = "10.18653/v1/2023.semeval-1.314",
    pages = "2304--2318"
}

@article{paun-etal-2018,
    title = "Comparing {B}ayesian Models of Annotation",
    author = "Paun, Silviu  and
      Carpenter, Bob  and
      Chamberlain, Jon  and
      Hovy, Dirk  and
      Kruschwitz, Udo  and
      Poesio, Massimo",
    journal = "Transactions of the Association for Computational Linguistics",
    volume = "6",
    year = "2018",
    address = "Cambridge, MA",
    publisher = "MIT Press",
    url = "https://aclanthology.org/Q18-1040/",
    doi = "10.1162/tacl_a_00040",
    pages = "571--585"
}

@inproceedings{luo-etal-2020-detecting,
    title = "Detecting Stance in Media On Global Warming",
    author = "Luo, Yiwei  and
      Card, Dallas  and
      Jurafsky, Dan",
    booktitle = "Findings of the Association for Computational Linguistics: EMNLP 2020",
    month = nov,
    year = "2020",
    address = "Online",
    publisher = "Association for Computational Linguistics",
    url = "https://aclanthology.org/2020.findings-emnlp.296/",
    doi = "10.18653/v1/2020.findings-emnlp.296",
    pages = "3296--3315"
}

@inproceedings{yin2023annobert,
  title={Annobert: Effectively representing multiple annotators’ label choices to improve hate speech detection},
  author={Yin, Wenjie and Agarwal, Vibhor and Jiang, Aiqi and Zubiaga, Arkaitz and Sastry, Nishanth},
  booktitle={Proceedings of the International AAAI Conference on Web and Social Media},
  volume={17},
  pages={902--913},
url={https://arxiv.org/abs/2212.10405},
  year={2023}
}

@inproceedings{fell2021mining,
  title={Mining Annotator Perspectives from Hate Speech Corpora.},
  author={Fell, Michael and Akhtar, Sohail and Basile, Valerio},
  booktitle={NL4AI@ AI* IA},
  url={https://ceur-ws.org/Vol-3015/paper136.pdf},
  year={2021}
}

@inproceedings{alabi-etal-2022-adapting,
    title = "Adapting Pre-trained Language Models to {A}frican Languages via Multilingual Adaptive Fine-Tuning",
    author = "Alabi, Jesujoba O.  and Adelani, David Ifeoluwa  and
      Mosbach, Marius  and Klakow, Dietrich",
    booktitle = "Proceedings of the 29th International Conference on Computational Linguistics",
    month = oct,
    year = "2022",
    address = "Gyeongju, Republic of Korea",
    publisher = "International Committee on Computational Linguistics",
    url = "https://aclanthology.org/2022.coling-1.382/",
    pages = "4336--4349"
}

@inproceedings{g-etal-2025-overview,
    title = "Overview of the Shared Task on Multimodal Hate Speech Detection in {D}ravidian languages: {D}ravidian{L}ang{T}ech@{NAACL} 2025",
    author = "G, Jyothish Lal  and B, Premjith  and Chakravarthi, Bharathi Raja  and
      Rajiakodi, Saranya  and B, Bharathi  and Natarajan, Rajeswari  and
      Rajalakshmi, Ratnavel",
    booktitle = "Proceedings of the Fifth Workshop on Speech, Vision, and Language Technologies for Dravidian Languages",
    month = may,
    year = "2025",
    address = "Acoma, The Albuquerque Convention Center, Albuquerque, New Mexico",
    publisher = "Association for Computational Linguistics",
    url = "https://aclanthology.org/2025.dravidianlangtech-1.20/",
    pages = "114--122",
    ISBN = "979-8-89176-228-2"
}

@article{braun2024beg,
  title={I beg to differ: how disagreement is handled in the annotation of legal machine learning data sets},
  author={Braun, Daniel},
  journal={Artificial intelligence and law},
  volume={32},
  number={3},
  pages={839--862},
  year={2024},
 url = {https://link.springer.com/article/10.1007/s10506-023-09369-4},
  publisher={Springer}
}

@inproceedings{barbieri-etal-2022-xlm,
    title = "{XLM}-{T}: Multilingual Language Models in {T}witter for Sentiment Analysis and Beyond",
    author = "Barbieri, Francesco  and Espinosa Anke, Luis  and   Camacho-Collados, Jose",
    booktitle = "Proceedings of the Thirteenth Language Resources and Evaluation Conference",
    month = jun,
    year = "2022",
    address = "Marseille, France",
    publisher = "European Language Resources Association",
    url = "https://aclanthology.org/2022.lrec-1.27/",
    pages = "258--266"
}

@misc{conneau2020unsupervised,
      title={Unsupervised Cross-lingual Representation Learning at Scale}, 
      author={Alexis Conneau and Kartikay Khandelwal and Naman Goyal and Vishrav Chaudhary and Guillaume Wenzek and Francisco Guzmán and Edouard Grave and Myle Ott and Luke Zettlemoyer and Veselin Stoyanov},
      year={2020},
      eprint={1911.02116},
      archivePrefix={arXiv},
      primaryClass={cs.CL},
      url={https://arxiv.org/abs/1911.02116}, 
}

@inproceedings{parmar-etal-2023-dont,
    title = "Don`t Blame the Annotator: Bias Already Starts in the Annotation Instructions",
    author = "Parmar, Mihir  and
      Mishra, Swaroop  and
      Geva, Mor  and
      Baral, Chitta",
    booktitle = "Proceedings of the 17th Conference of the European Chapter of the Association for Computational Linguistics",
    month = may,
    year = "2023",
    address = "Dubrovnik, Croatia",
    publisher = "Association for Computational Linguistics",
    url = "https://aclanthology.org/2023.eacl-main.130/",
    doi = "10.18653/v1/2023.eacl-main.130",
    pages = "1779--1789"
}

@inproceedings{muhammad-etal-2022-naijasenti,
    title = "{N}aija{S}enti: A {N}igerian {T}witter Sentiment Corpus for Multilingual Sentiment Analysis",
    author = "Muhammad, Shamsuddeen Hassan  and  Adelani, David Ifeoluwa  and
      Ruder, Sebastian  and  Ahmad, Ibrahim Sa{'}id  and Abdulmumin, Idris  and
      Bello, Bello Shehu  and Choudhury, Monojit  and Emezue, Chris Chinenye  and
      Abdullahi, Saheed Salahudeen  and Aremu, Anuoluwapo  and Jorge, Al{\'i}pio  and
      Brazdil, Pavel",
    booktitle = "Proceedings of the Thirteenth Language Resources and Evaluation Conference",
    month = jun,
    year = "2022",
    address = "Marseille, France",
    publisher = "European Language Resources Association",
    url = "https://aclanthology.org/2022.lrec-1.63/",
    pages = "590--602"
}

@article{jiang-marneffe-2022-investigating,
    title = "Investigating Reasons for Disagreement in Natural Language Inference",
    author = "Jiang, Nan-Jiang  and
      de Marneffe, Marie-Catherine",
    journal = "Transactions of the Association for Computational Linguistics",
    volume = "10",
    year = "2022",
    address = "Cambridge, MA",
    publisher = "MIT Press",
    url = "https://aclanthology.org/2022.tacl-1.78/",
    doi = "10.1162/tacl_a_00523",
    pages = "1357--1374"
}

@article{Cabitza_2023,
   title={Toward a Perspectivist Turn in Ground Truthing for Predictive Computing},
   volume={37},
   ISSN={2159-5399},
   url={http://dx.doi.org/10.1609/aaai.v37i6.25840},
   DOI={10.1609/aaai.v37i6.25840},
   number={6},
   journal={Proceedings of the AAAI Conference on Artificial Intelligence},
   publisher={Association for the Advancement of Artificial Intelligence (AAAI)},
   author={Cabitza, Federico and Campagner, Andrea and Basile, Valerio},
   year={2023},
   month=jun, pages={6860–6868} }

@misc{rethinkingemotion2024,
      title={Rethinking Emotion Annotations in the Era of Large Language Models}, 
      author={Minxue Niu and Yara El-Tawil and Amrit Romana and Emily Mower Provost},
      year={2024},
      eprint={2412.07906},
      archivePrefix={arXiv},
      url={https://arxiv.org/abs/2412.07906}, 
}

@article{kappa,
 ISSN = {0006341X, 15410420},
 URL = {http://www.jstor.org/stable/2529310},
 author = {J. Richard Landis and Gary G. Koch},
 journal = {Biometrics},
 number = {1},
 pages = {159--174},
 publisher = {International Biometric Society},
 title = {The Measurement of Observer Agreement for Categorical Data},
 urldate = {2026-03-12},
 volume = {33},
 year = {1977}
}

@article{davani_dealing_2022,
    title = {Dealing with {Disagreements}: {Looking} {Beyond} the {Majority} {Vote} in {Subjective} {Annotations}},
    volume = {10},
    issn = {2307-387X},
    shorttitle = {Dealing with {Disagreements}},
    url = {https://doi.org/10.1162/tacl_a_00449},
    doi = {10.1162/tacl_a_00449},
    urldate = {2024-12-26},
    journal = {Transactions of the Association for Computational Linguistics},
    author = {Davani, Aida Mostafazadeh and Díaz, Mark and Prabhakaran, Vinodkumar},
    month = jan,
    year = {2022},
    pages = {92--110},
}
\bibliographystyle{acl_natbib}

\end{document}